\documentclass[journal]{IEEEtran}
\usepackage{graphicx}
\usepackage{amsfonts}

\usepackage[english]{babel}
\usepackage{amsmath, amssymb}
\usepackage{float}
\usepackage{makecell}
\usepackage{multirow}
\usepackage{cite}
\usepackage{amsmath,amssymb,amsfonts}
\usepackage{algorithmic}
\usepackage{graphicx}
\usepackage{textcomp}
\usepackage{xcolor}
\usepackage{subfigure}
\usepackage{epstopdf}
\usepackage[T1]{fontenc}
\usepackage{graphicx}
\usepackage{comment}
\usepackage{color}
\usepackage{enumerate}

\usepackage{booktabs}
\usepackage[utf8]{inputenc} 
\usepackage{hyperref}       
\usepackage{url}            
\usepackage{nicefrac}       
\usepackage{microtype}      
\usepackage{wrapfig}
\usepackage{epsfig}
\usepackage{stfloats}
\usepackage{float}
\usepackage{soul}
\soulregister\cite7 
\soulregister\citep7 
\soulregister\citet7 
\soulregister \ref7 
\soulregister\pageref7 
\soulregister\item7
\sethlcolor{yellow}
\newcommand\modify[1]{#1}
\newcommand\modifynew[1]{#1}
\newcommand\modifylatest[1]{#1}
\ifCLASSINFOpdf
\else
\fi
\begin{document}
%
\title{End-to-end 2D-3D Registration between Image and LiDAR Point Cloud for Vehicle Localization}
%
%
%

\author{Guangming~Wang, Yu Zheng, Yuxuan Wu,
        Yanfeng Guo, Zhe Liu, Yixiang Zhu, Wolfram Burgard, \\ and Hesheng Wang
        
\thanks{*This work was supported in part by the Natural Science Foundation of China under Grant 62225309, Grant U24A20278, Grant
62361166632, and Grant U21A20480. The first two authors contributed equally. Corresponding Author: Hesheng Wang.}
\thanks{G. Wang is with the Department of Engineering, University of Cambridge, Cambridge CB2 1PZ, U.K.}

\thanks{Y. Zheng, Y. Wu, Z. Liu, and H. Wang are with the School of Automation and Intelligent Sensing, Shanghai Jiao Tong University, Shanghai 200240, Chia and Key Laboratory of System Control and Information Processing, Ministry of Education of China, Shanghai 200240, China.}
\thanks{Y. Guo is with Electrical and Computer
Engineering at the University of California, Los Angeles,
the United States.}
\thanks{Y. Zhu is with Computer Control and Automation at Nanyang Technological University, Singapore.}
\thanks{W. Burgard is with the Department of Computer Science and Artificial Intelligence, University of Technology Nuremberg, Germany.}
}

%
%

\markboth{Journal of \LaTeX\ Class Files,~Vol.~14, No.~8, August~2015}%
{Shell \MakeLowercase{\textit{et al.}}: Bare Demo of IEEEtran.cls for IEEE Journals}
%



\maketitle
\begin{abstract}
Robot localization using a built map is essential for a variety of tasks including accurate navigation and mobile manipulation. A popular approach to robot localization is based on image-to-point cloud registration, which combines illumination-invariant LiDAR-based mapping with economical image-based localization. However, the recent works for image-to-point cloud registration either divide the registration into separate modules or project the point cloud to the depth image to register the RGB and depth images. \modifynew{In this paper, we present I2PNet, a novel end-to-end 2D-3D registration network, which directly registers the raw 3D point cloud with the 2D RGB image using differential modules with a united target.} \modifynew{The 2D-3D cost volume module for differential 2D-3D association is proposed to bridge feature extraction and pose regression. The soft point-to-pixel correspondence is implicitly constructed on the intrinsic-independent normalized plane in the 2D-3D cost volume module.} Moreover, we introduce an outlier mask prediction module to filter the outliers in the 2D-3D association before pose regression. Furthermore, we propose the coarse-to-fine 2D-3D registration architecture to increase localization accuracy.  Extensive localization experiments are conducted on the KITTI, nuScenes, M2DGR, Argoverse, Waymo, and Lyft5 datasets. The results demonstrate that I2PNet outperforms the state-of-the-art by a large margin and \modifylatest{has a higher efficiency than the previous works.} Moreover, we extend the application of I2PNet to the camera-LiDAR online calibration and demonstrate that I2PNet outperforms recent approaches on the online calibration task. \modify{Source codes are released at https://github.com/IRMVLab/I2PNet.}
\end{abstract}


%
\IEEEpeerreviewmaketitle

\section{Introduction}
%
%
%
%

\IEEEPARstart{h}{igh-accuracy} robot localization in pre-built maps is an essential task for autonomous mobile robots, enabling high-accuracy robot navigation and mobile manipulation. \modifynew{Recently, most researches focus on same-modality mapping and localization based on Light Detection And Ranging (LiDAR) point cloud or images.} The LiDAR point cloud-based mapping and localization can realize considerable localization accuracy with the place recognization and point cloud-to-point cloud registration~\cite{besl1992method,yang2013go,liu2019lpd,lu2019deepvcp,yew2020rpm,cattaneo2022lcdnet}. However, this localization method requires the mobile robot equipped with the expensive LiDAR, which greatly limits the popularity of mobile robots. Meanwhile, the accuracy of image-based mapping and localization~\cite{lowe1999object,ke2004pca,sattler2011fast,sattler2016efficient,yi2016lift} is greatly limited in poor illumination and featureless environments. \modifynew{In contrast to same-modality method, when using cross-modality mapping and localization, i.e., LiDAR-based mapping and monocular camera-based localization, the robot is only required to be equipped with the economical monocular camera.} Additionally, the LiDAR point cloud-based mapping is invariant to the illumination and can represent the 3D scenes with high accuracy even in featureless environments. Therefore, cross-modality mapping and localization are more promising than same-modality mapping and localization and worthy to research.

\begin{figure}[t]
	\centering
	\vspace{0mm}
	\resizebox{0.8\linewidth}{!}
	{
		\includegraphics[scale=1.0]{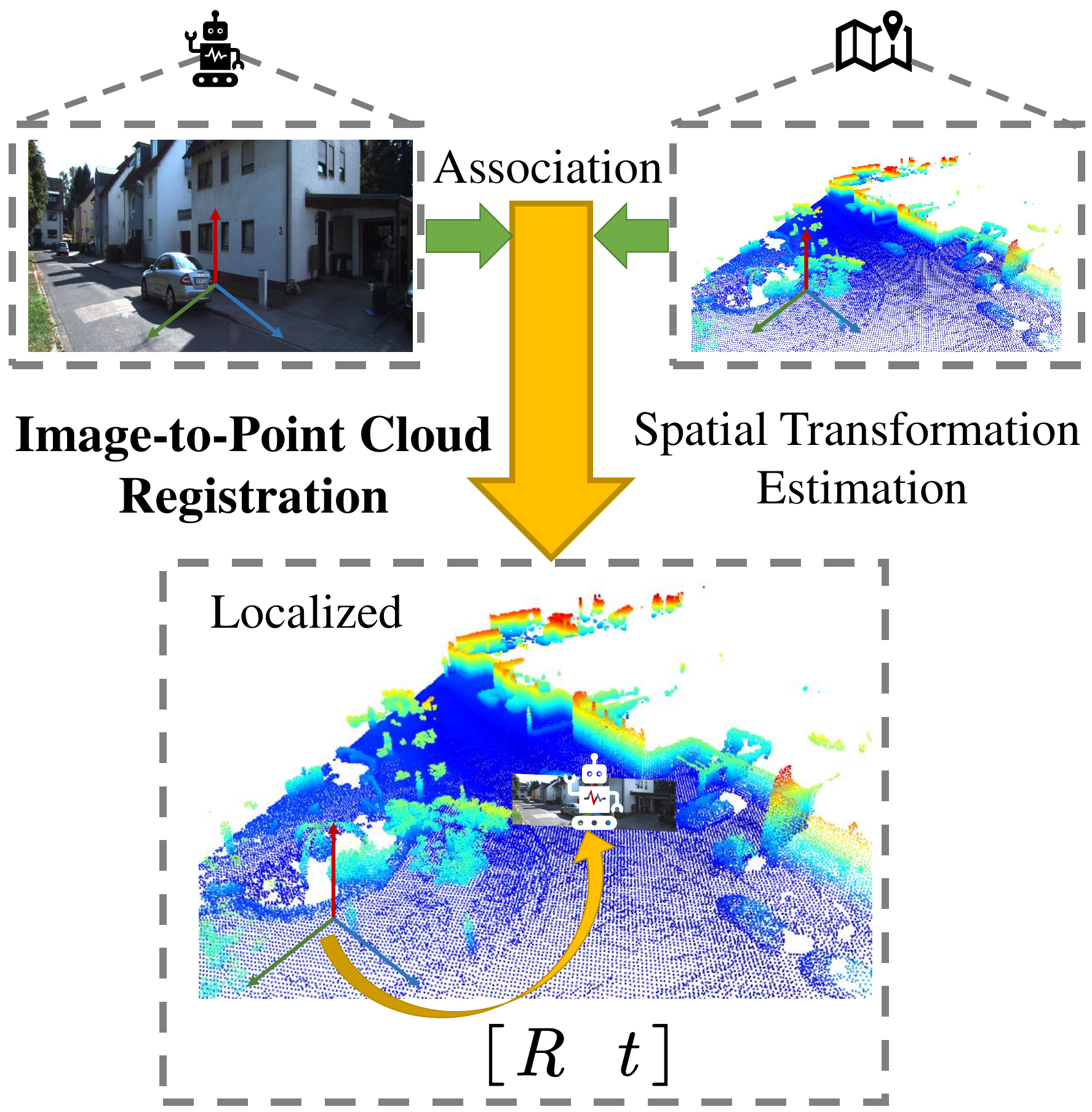}}
	\vspace{-2mm}
	\caption{ \modifynew{The pipeline of monocular camera localization in LiDAR maps with image-to-point cloud registration.} Image-to-point cloud registration associates the image and point cloud and estimates the spatial transformation between them. With the estimated transformation, the pose of the monocular camera in the LiDAR map is obtained and the robot is localized.}
	\vspace{-3mm}
	\label{fig:prop}
\end{figure}

The conventional monocular camera localization in the LiDAR map adopts the matching between the camera image and synthetic image from LiDAR map~\cite{wolcott2014visual,pascoe2015direct}, or registration between LiDAR map and local 3D point cloud reconstructed from the image sequence~\cite{caselitz2016monocular}. \modifynew{However, the rendering of synthetic images is costly.} In addition, the 3D local reconstruction requires image sequence input and can fail in featureless scenes. Therefore, image-to-point cloud registration, which directly registers the image and LiDAR point cloud as shown in Fig.~\ref{fig:prop}, is more suitable for monocular camera localization in the  LiDAR point cloud map. However, image-to-point cloud registration is more complex than same-modality registration due to the different characteristics of the different modalities. The RGB image contains rich texture information which the LiDAR point cloud lacks, while the LiDAR point cloud contains rich 3D geometric information which the RGB image lacks.
Therefore, image-to-point cloud registration for vehicle localization starts to be widely researched after the occurrence of brain-inspired~\cite{qiao2023brain} Deep Neural Networks (DNNs). Most DNN-based image-to-point cloud registration methods for vehicle localization divide the image-to-point cloud registration into separate modules~\cite{feng20192d3d,wang2021p2,ren2022corri2p,li2021deepi2p,jeon2022efghnet,cattaneo2020cmrnet++}. DNN-based methods can improve the performance of a part of the modules, such as image-to-point cloud association~\cite{feng20192d3d,wang2021p2,ren2022corri2p,li2021deepi2p}. 
However, the separate modules are designed or optimized for separate targets. The error of the former modules will not be corrected in the subsequent modules since they are not jointly optimized for the united target. 
In addition, several works~\cite{feng20192d3d,wang2021p2,ren2022corri2p,li2021deepi2p,cattaneo2020cmrnet++} adopt the Random SAmple Consensus (RANSAC)~\cite{fischler1981random} based Perspective-n-Point (PnP) solvers~\cite{abdel2015direct,lepetit2009epnp} or non-linear optimization solvers~\cite{marquardt1963algorithm} as one of the separate modules. These iterative modules limit the efficiency of these methods.
Recently, several works~\cite{cattaneo2019cmrnet,chang2021hypermap,chen2022i2d} attempt to register the image with point cloud in an end-to-end manner. However, these works project the raw point cloud into a depth image, utilize Convolutional Neural Networks (CNNs) to associate the RGB image and depth image, and finally regress the spatial transformation. The projection of the point cloud limits the application of these methods to large-range localization, since a great number of points lose during the projection when the misalignment between the point cloud and the image is large.

In this paper, we introduce I2PNet, a novel end-to-end image-to-point cloud registration method for vehicle localization. In contrast to recent methods, I2PNet is a fully end-to-end architecture without separate modules and directly associates the 2D RGB image and the raw 3D LiDAR point cloud. To realize end-to-end 2D-3D registration, three challenges need to be overcome: 1) Since the 3D coordinates of LiDAR points are contiguous but the 2D coordinates of the image pixels are discrete, a LiDAR point can hardly find a precisely corresponding pixel;
2) To make the model not limited by a specific camera, the image-to-point cloud association should be independent of the camera intrinsic; 
3) Since the LiDAR point cloud fully covers the surrounding area while the image only covers the front area, many point-to-pixel associations are outliers and should be automatically filtered in a differential manner. 

\modifynew{For the first two challenges, we propose a novel differential 2D-3D association module, named 2D-3D cost volume module, to achieve local feature association between the 3D point cloud and the 2D image on a camera intrinsic-independent normalized image plane. Feature association is realized by querying pixel features around each projected 3D point in the intrinsic-independent normalized image plane and performing feature similarity calculations to generate differentiable implicit correspondences. This differential 2D-3D cost volume module bridges feature extraction and pose regression, enabling end-to-end registration between the 3D point cloud and the 2D image. Therefore,  we solve the first challenge by generating point-wise implicit point-to-pixel correspondences in the 2D-3D cost volume module and enabling all the points to be softly associated with the pixels. We solve the second challenge by performing the image-to-point cloud association on the normalized plane of the pinhole camera model since the coordinates on the normalized plane are naturally independent of the camera intrinsic.}

For the third challenge, an outlier mask prediction module is proposed. In this module, the 2D-3D cost volumes and LiDAR point features are utilized to generate outlier masks. For more accurate outlier filtering, the context features of point-to-pixel implicit correspondence are gathered to embed the patch spatial transformation information in the 2D-3D cost volumes.
The geometric transformation information improves the outlier prediction and results in better registration. Finally, the 2D-3D cost volumes are masked by the outlier masks and aggregated to estimate the spatial transformation between the point cloud and the image. 
In addition, we propose a coarse-to-fine 2D-3D registration architecture and perform fine
registration based on the coarse prior knowledge. The coarse registration results are used to warp the point cloud to gain a pair of the image and point cloud with smaller misalignment for the fine registration. Furthermore, the coarse cost volumes are fused with the fine cost volumes to gain a better spatial
transformation estimation.

In summary, our main contributions are:
\modifylatest{
\begin{itemize}
\item 
We introduce a novel end-to-end 2D-3D registration architecture, named I2PNet, for vehicle localization. Different from existing methods, all the modules in our architecture are jointly optimized by a united target, and the complete 3D point cloud is preserved for large-range localization.
\item We propose the novel 2D-3D cost volume module to enable end-to-end 2D-3D registration. 
2D-3D cost volume module differentially associates 3D points and 2D pixels on the camera intrinsic-independent space.
\item We conduct extensive robot localization experiments on KITTI~\cite{geiger2013vision}, nuScenes~\cite{nuscenes2019}, M2DGR~\cite{yin2021m2dgr}, Argoverse~\cite{chang2019argoverse}, Waymo~\cite{sun2020waymo}, and Lyft5~\cite{houston2021one} datasets and various localization ranges to show the superiority and generalization of I2PNet. Moreover, we evaluate the efficiency of I2PNet and demonstrate the end-to-end pipeline can improve both performance and efficiency. 
\item We extend the application of I2PNet to the camera-LiDAR online calibration and demonstrate the effectiveness of I2PNet on various tasks.
\end{itemize}
}

\section{Related Work}

\modifylatest{In this section, we will introduce the state-of-the-art image-to-point cloud registration works for robot localization, followed by a review of state-of-the-art camera-LiDAR online calibration methods.}

\subsection{\modifylatest{Image-to-Point Cloud Registration for Robot Localization}}\label{sec:cross modality rw}
\modifylatest{
2D3D-MatchNet~\cite{feng20192d3d} is one of the earliest works focusing on image-to-point cloud registration for robot localization. 
The 2D and 3D keypoints are obtained by SIFT and ISS~\cite{zhong2009intrinsic} respectively. Then, a neural network with three branches is introduced to learn the descriptors for keypoints. Finally, EPnP~\cite{lepetit2009epnp} is adopted to estimate the transformation between the image and the point cloud with the 2D-3D correspondences.
DeepI2P~\cite{li2021deepi2p} splits the image-to-point cloud registration into a classification problem and an optimization problem. A cross-modality neural network is adopted to classify whether the points fall into the image frustum. The classification results are utilized to construct the cost function of the inverse camera projection. The optimization solver solves the transformation that minimizes the value of the cost function.
CorrI2P~\cite{ren2022corri2p} designs a cross-modality network to extract the image-to-point cloud overlapping region and corresponding dense descriptors for the image and point cloud. CorrI2P constructs dense image-to-point cloud correspondences and uses iterative RANSAC-based EPnP~\cite{lepetit2009epnp} to estimate the relative pose. 
EFGHNet~\cite{jeon2022efghnet} adopts the divide-and-conquer strategy to divide the image-to-point cloud registration into four separate sub-networks. These sub-networks are responsible for the horizon and ground normal alignments, rotation estimation, and translation estimation. The four sub-networks are sequentially applied and the subsequent networks depend on the results of the previous networks. These methods divide the image-to-point cloud registration into separate modules for large-range robot localization. The separation makes the modules separately optimized and thus not able to refine the error of the previous modules.
}

\modifylatest{
In addition, inspired by the coarse-to-fine architecture adopted in the fields including optical flow estimation~\cite{sun2018pwc}, scene flow estimation~\cite{wu2020pointpwc,wang2021hierarchical,wang2022matters}, and deep LiDAR odometry~\cite{wang2021pwclo}, a few recent works~\cite{cattaneo2019cmrnet, chang2021hypermap,chen2022i2d} attempt to form an end-to-end image-to-point cloud registration network for robot localization with the 2D-2D coarse-to-fine architecture~\cite{sun2018pwc}. CMRNet~\cite{cattaneo2019cmrnet} is one of the representative methods. 
CMRNet projects the point cloud as a depth image. Based on the depth image, it utilizes the CNN-based PWC-Net to perform the 2D-2D coarse-to-fine registration between the RGB and depth images.
CMRNet shows the effectiveness of the 2D-2D coarse-to-fine architecture in small-range localization. However, projecting point cloud as a depth image limits the application of CMRNet to large-range localization, since many points are dropped during the projection when the misalignment between image and point cloud is large. 
}

\modifylatest{
\textit{In I2PNet, all parts are differentially united and jointly optimized, which enables the error refinement of the subsequent modules and makes the registration more robust. Meanwhile, the end-to-end 2D-3D registration architecture avoids utilizing the depth image projected from point cloud and thus enables the application of I2PNet for large-range localization by preserving the complete 3D point cloud.}
}

\subsection{\modifylatest{Camera-LiDAR Online Calibration}}
\modifylatest{
The camera-LiDAR online calibration task is to online correct the calibration error between the camera and LiDAR. The conventional camera-LiDAR online calibration methods extract the common low-level features of the image and
point cloud, such as contours~\cite{bileschi2009fully,levinson2013automatic} or itensity~\cite{pandey2015automatic}, and match the features. They utilize feature matches to construct the cost function and optimize the cost function to gain the decalibration matrix. The first deep-learning-based online camera-LiDAR calibration method is RegNet~\cite{schneider2017regnet}. It utilizes several Network-In-Network (NIN) blocks~\cite{lin2013network} to extract the features of the RGB image and depth map to obtain image features and depth features respectively. 
The extracted image features and depth features are simply concatenated and fed into several NIN blocks and Fully Connected (FC) layers to perform feature matching and pose regression respectively. 
The subsequent works of RegNet focus on better loss functions to improve the calibration. CalibNet~\cite{iyer2018calibnet} introduces the photometric loss and point cloud distance loss to perform the geometrical supervision.  In addition,
RGGNet~\cite{yuan2020rggnet} introduces the geodesic distance loss to supervise the calibration in $se3$ space based on the Riemannian geometry and the tolerance regularizer loss to supervise the error bound with an implicit tolerance model. 
}

\modifylatest{
The recent DNN-based camera-LiDAR calibration mostly adopts the depth image representation of the LiDAR point cloud and utilizes CNNs to realize the image-to-point cloud registration like CMRNet. Therefore, their application is limited in the online calibration task where the misalignment between the image and point cloud is small. }

\modifylatest{
\textit{The 2D-3D registration architecture of our I2PNet is suitable for both robot localization and online calibration. Moreover, I2PNet outperforms the recent online calibration methods.}   
}

\begin{figure*}[t]
	\centering
	\vspace{0mm}
	\resizebox{1.0\textwidth}{!}
	{
		\includegraphics[scale=1.0]{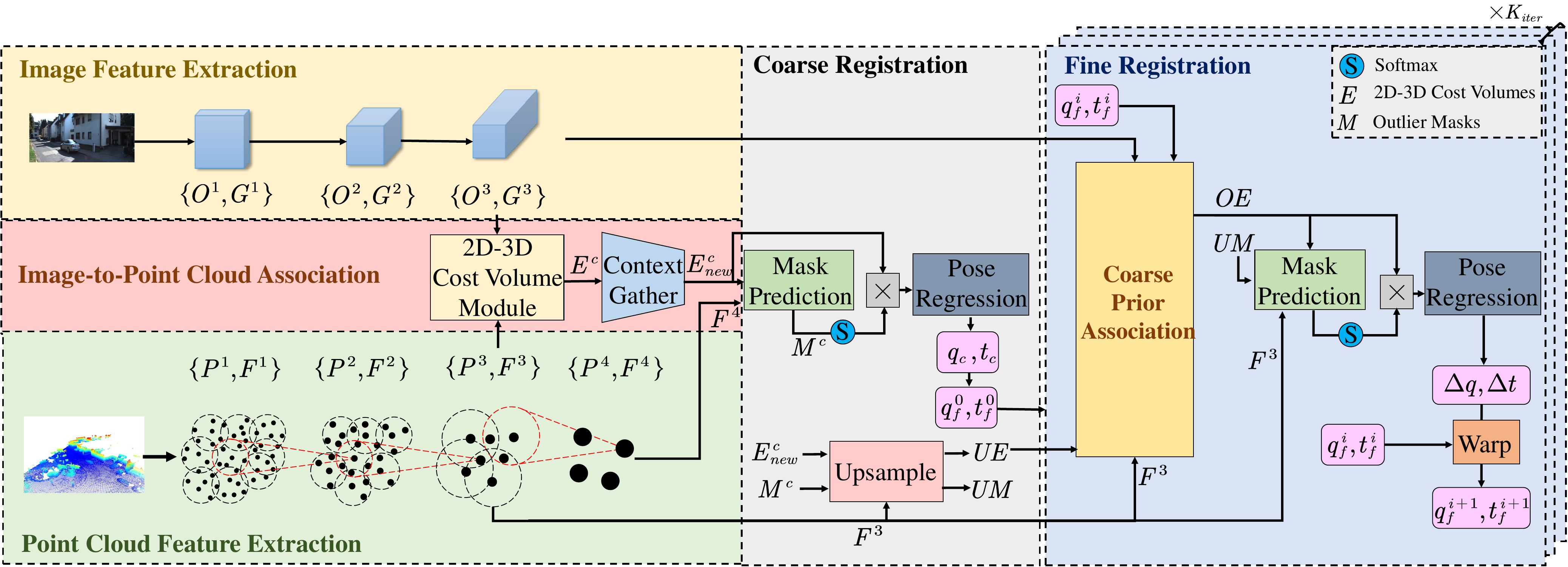}}
	\vspace{-5mm}
	\caption{\modifylatest{The outline of I2PNet. I2PNet takes the RGB image and the raw LiDAR point cloud as input. Through the feature extraction pyramid, coarse registration, and fine registration, the network finally predicts relative pose between the image and point cloud. The detailed structure of coarse prior association is shown in Fig.~\ref{fig:msup}.}
	}
	\vspace{-4mm}
	\label{fig:ms}
\end{figure*}
\section{End-to-end 2D-3D Registration}
\subsection{Network Architecture}\label{section:overview}
Image-to-point cloud registration is defined as that given the RGB image and the point cloud, the network estimates the spatial transformation between the image and the point cloud. 
I2PNet performs end-to-end 2D-3D image-to-point cloud registration with three main components: feature extraction, coarse registration, and fine registration. The main architecture of I2PNet is shown in Fig.~\ref{fig:ms}.

In the feature extraction, the image and point cloud feature pyramids extract the hierarchical image and point features for the image-to-point cloud association. 

In the coarse registration, the 2D-3D cost volume module associates the image and point cloud in the third layer of the pyramid and outputs the 2D-3D cost volumes. The context gathering module further aggregates the 2D-3D cost volumes. Then, the outlier masks are predicted from the 2D-3D cost volumes and point features. The 2D-3D cost volumes are masked by the outlier masks for the outlier filtering and pose regression. \modifylatest{The regressed coarse relative pose is treated as the initial pose for the fine registration. In addition, the 2D-3D cost volumes and outlier masks are upsampled for the fine registration.}

\begin{figure}[t]
	\centering
	\vspace{0mm}
	\resizebox{0.8\linewidth}{!}
	{
		\includegraphics[scale=1.0]{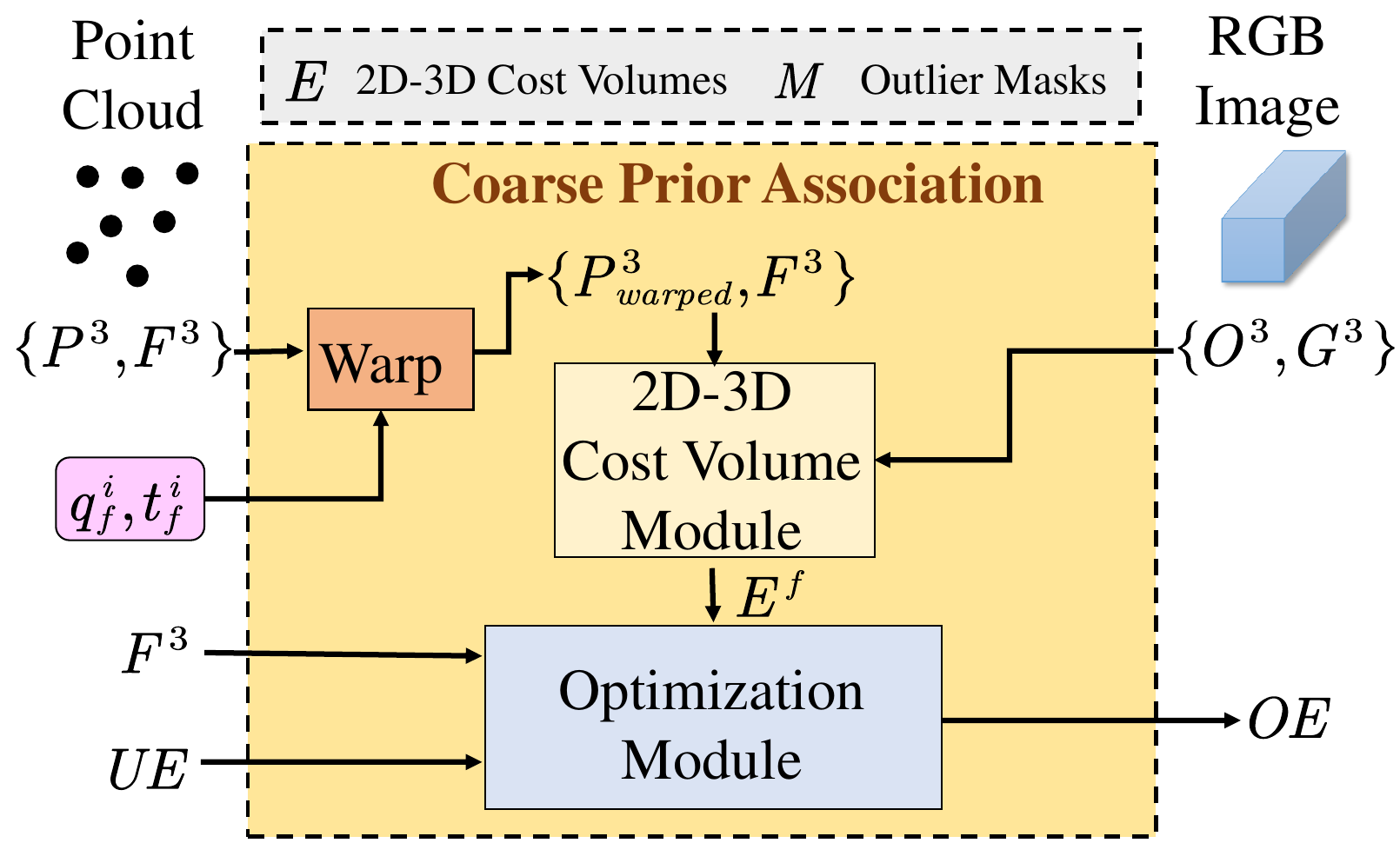}}
	\vspace{-2mm}
	\caption{\modifylatest{The detailed structure of the coarse prior association in Fig.~\ref{fig:ms}.}}
	\label{fig:msup}
 	\vspace{-5mm}
\end{figure}

\modifylatest{In the fine registration, we propose the coarse prior association module to transfer the prior knowledge from the coarse registration to the fine registration. In the coarse prior association, as shown in Fig.~\ref{fig:msup}, 
the regressed coarse relative pose is used to warp the point cloud.} After the warping, the warped point cloud and the image are associated to generate the 2D-3D cost volumes of the residual spatial transformation, which are then fused with the upsampled 2D-3D cost volumes in the optimization module to output the optimized 2D-3D cost volumes.

The subsequent modules in fine registration estimate the residual transformation based on the results of coarse prior association, as shown in Fig.~\ref{fig:ms}. Specifically, the fine outlier masks are predicted using the optimized 2D-3D cost volumes and fused with the upsampled outlier masks to gain the optimized outlier masks. The optimized 2D-3D cost volumes are masked by the optimized outlier mask and regress the residual relative pose through the pose regression.
\modifylatest{The coarse relative pose is warped by the residual relative pose to obtain the refined relative pose. The fine registration is an iterative architecture. For the following iteration process, the relative pose predicted in the last iteration is used to warp the point cloud and refined by the residual relative pose predicted in the iteration. The iterations are conducted $K_{iter}$ times, and the estimated relative pose in the final iteration is output as the fine relative pose.}

\subsection{Feature Extraction}\label{section:fe}
\subsubsection{Image Feature Extraction} The features of the input RGB image are extracted by three layers. Each layer consists of $5$ convolutional blocks. Each convolutional block is the composition of $3 \times 3$ convolution, batch normalization, leaky Rectified-Linear Unit (ReLU), and max-pooling. Max-pooling is adopted to downsample the image to different resolutions. The extracted image features in the $l$-th layer are $G^l=\{{g}_{i}^l|i=1,2,...,M^l\}$, in which $M^l$ is the number of pixels. The 2D coordinates of the pixels on the pixel plane are represented by the position information ${O^l}=\{{{o}_{i}^l}|i=1,2,...,M^l\}$.

\subsubsection{Point Cloud Feature Extraction} The features of the input point cloud are extracted by PointNet++~\cite{qi2017pointnet++}. PointNet++ adopts Farthest Point Sampling (FPS) to select an evenly distributed subset of the original point cloud. Each point in the selected subset is treated as the center point of a point group. For each center point, several points are grouped by querying the nearest neighbors from all the points in the point cloud. The features of each point group are aggregated by PointNet~\cite{qi2017pointnet}.
However, because of the high time complexity of FPS and neighborhood query among all the points, the vanilla PointNet++ is inefficient for the large-scale point cloud from the LiDAR point cloud map. Therefore, we refer to EfficientLO-Net~\cite{wang2021efficient} to use the stride-based sampling and projection-aware grouping to replace the sampling and neighborhood query methods in PointNet++. To apply the stride-based sampling and projection-aware grouping, the 2D spherical coordinates $(u_s,v_s)$ of each point are calculated as~\cite{wu2018squeezeseg}:
\begin{equation}
    \left(\begin{matrix}u_s\\v_s\end{matrix}\right) = \left(\begin{matrix}\frac{1}{2}\lfloor1 - \arctan(y,x)\cdot\pi^{-1}\rfloor \cdot W\\ \lfloor1-(\arcsin(z / r)+ f_{down})\cdot f^{-1}\rfloor\cdot H\end{matrix}\right),
\end{equation}
in which $(x,y,z)$ are the 3D coordinates of each point, and $r=\sqrt{x^2+y^2+z^2}$ is the range of each point. $f=f_{up}+f_{down}$ is the vertical field-of-view of the LiDAR sensor, in which $f_{up}$ and $f_{down}$ are the up and down vertical field-of-view respectively. $H$ and $W$ are the initial upper bounds of the 2D spherical coordinates, i.e., $0\le u_s\textless W$ and $0\le v_s \textless H$. Based on the 2D spherical coordinates, stride-based sampling can perform efficient sampling to obtain the center points. The stride-based sampling refers to the stride mechanism of the 2D convolution. The points whose 2D coordinates are exactly integral multiples of the strides are sampled as the center points. After obtaining the center points, projection-aware grouping performs the efficient neighborhood query by reducing the search space of the 3D nearest neighbors. A fixed-size 2D kernel whose center is the 2D coordinates of each sampled center point is adopted for the searching space reduction. Therefore, the searching space is reduced from the whole point cloud to the kernel points whose 2D coordinates are inside the fixed-size 2D kernel.  
Due to the characteristic of spherical projection, the 3D nearest neighbors are included in the kernel points when selecting an appropriate kernel size~\cite{wang2021efficient}. Therefore, the 3D nearest neighbors are queried from the kernel points through the KNN algorithm.
It is noticed that the points whose distances towards the center point are over the distance threshold will not be selected by KNN. Overall, by stride-based sampling and projection-aware grouping, I2PNet can efficiently extract features of the large-scale raw point cloud.

The point features are extracted by four layers from fine-grained to coarse-grained in point cloud feature extraction. 
In the $l$-th layer, $P^{l}=\{{p}_{i}^l|i=1,2,...,N^l\}$ represent the position information, and $F^{l}=\{{f}_{i}^l|i=1,2,...,N^l\}$ represent the point features, where $N^l$ is the number of the points. Notably, $P^{0}$ are the coordinates of the input point cloud while $F^{0}$ are the initial point features. The feature extraction is as follows:
\begin{equation}
	f_i^l = MaxPool(MLP(\{f_{i,k}^{l-1}\}_{k=1}^{K^l})),
\end{equation}
where $f_{i,k}^{l-1}$ is the feature of $k$-th nearest neighbors and $K^l$ is the number of points in a point group. $MaxPool$ means max-pooling operation. $MLP$ means the shared MLP block. In addition, the position information of the next layer is obtained as $P^l = \mathcal{S}(P^{l-1})$, where $\mathcal{S}$ is the sampling method.

\begin{figure*}[t]
	\centering
	\vspace{0mm}
	\resizebox{1.0\textwidth}{!}
	{
		\includegraphics[scale=1.0]{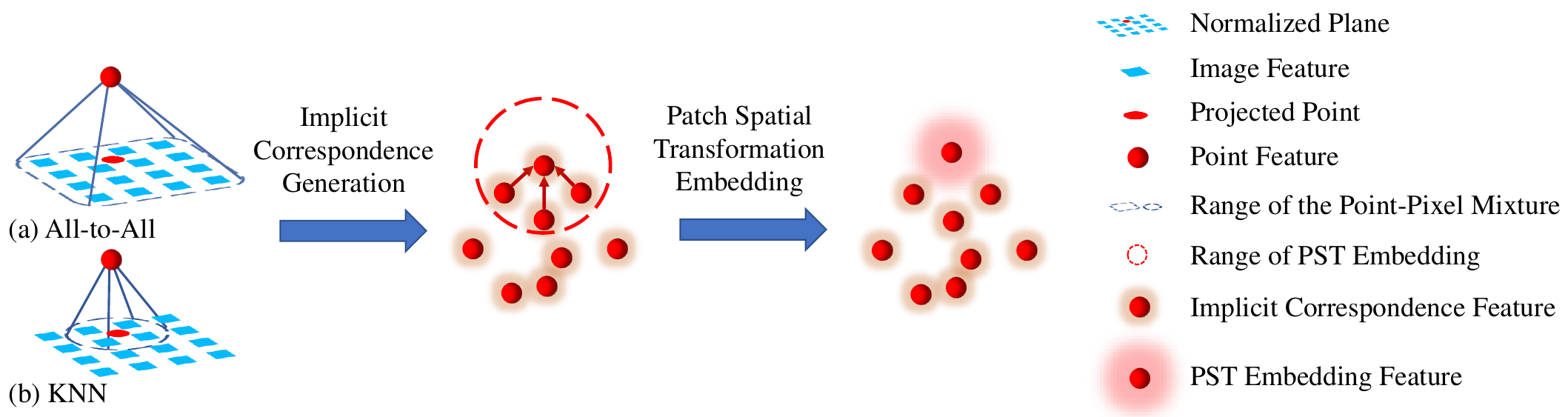}}
	\vspace{-6mm}
	\caption{\modify{The process diagram of the 2D-3D cost volume module. The Implicit Correspondence (IC) generation module uses an all-to-all or KNN-based point-pixel mixture to match the points and pixels on the normalized plane of the pinhole camera model. Then, the similarities of each point-pixel pair are calculated and aggregated to generate the point-wise IC features. Based on the IC features, the Patch Spatial Transformation (PST) embedding module aggregates the IC features of the neighbors of each point to estimate the patch spatial transformation information and embeds it in the PST embedding features. The module finally outputs the PST embedding features as the 2D-3D cost volumes.}}
	\vspace{-2mm}
	\label{fig:cvow}
\end{figure*}

\begin{figure*}[t]
	\centering
	\vspace{0mm}
	\resizebox{1.0\textwidth}{!}
	{
        \includegraphics[scale=1.0]{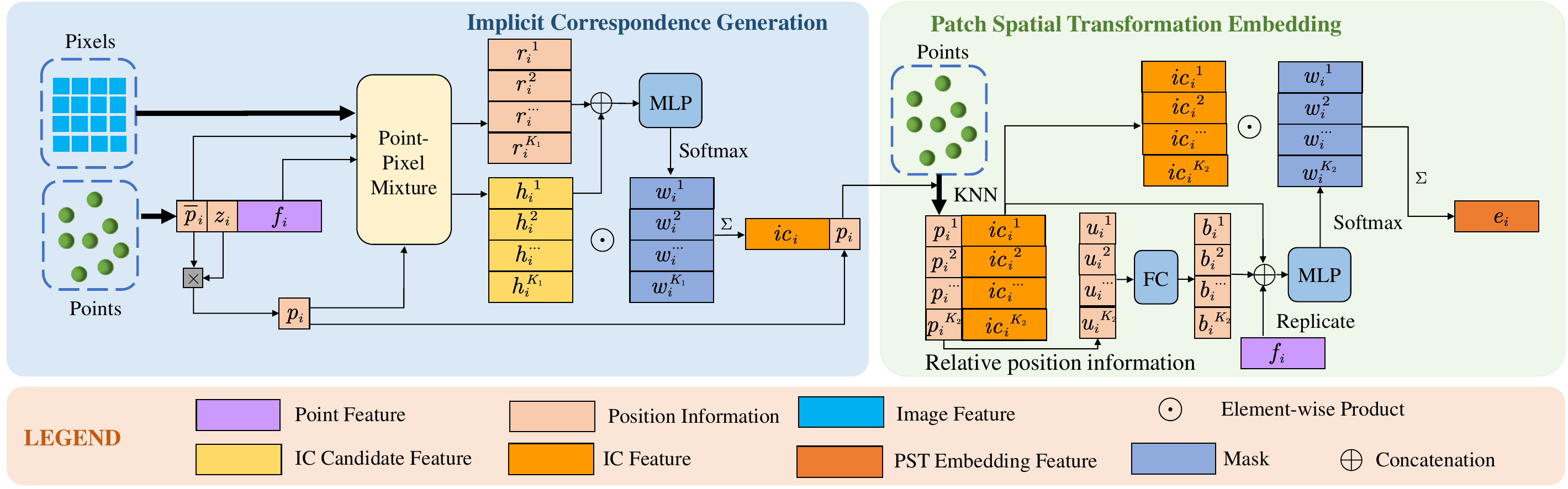}}
	\vspace{-6mm}
	\caption{\modify{The detailed pipeline of the 2D-3D cost volume module. Given the features of the image and point cloud, as well as their position information, the module estimates the PST embedding features as the 2D-3D cost volumes. The detailed structures of the point-pixel mixtures are shown in Fig. \ref{fig:cv2} and Fig. \ref{fig:cv1}.}}
	\vspace{-2mm}
	\label{fig:cvm}
\end{figure*}

\subsection{2D-3D Cost Volume Module}\label{section:2D-3D Cost Volume}
\subsubsection{Overview}\label{subsec:2D3DCVMoverview}

\modifynew{The 2D-3D cost volume module implicitly constructs the soft point-to-pixel correspondence on the intrinsic-independent normalized plane of the pinhole camera model. Specifically, the 2D-3D cost volume module first projects the 3D point cloud features onto the 2D normalized image plane, while simultaneously inverse-projecting the image features onto the 2D normalized plane using the camera intrinsics as shown in the left of Fig.~\ref{fig:cvow}. This process converts both the 3D points and image pixels into the same space, an intrinsic-independent normalized plane. On this plane, each 3D point queries the surrounding 2D pixel features and performs neighborhood feature aggregation. The aggregated features represent the matching relationships between each individual 3D point and the surrounding neighborhood of pixels. Since the correspondence information between individual 3D points and surrounding pixels is implicitly constructed as a feature, this process is referred to Implicit Correspondence (IC) generation module. In addition, the IC features are further gathered to embed the spatial transformation in raw 3D space in a Patch Spatial Transformation (PST) embedding module to generate final 2D-3D cost volumes.}

\subsubsection{Detailed Pipeline}
We present the detailed pipeline of the 2D-3D cost volume module in Fig.~\ref{fig:cvm}. In the implicit correspondence generation module, the point cloud and image are projected and inverse-projected respectively onto the camera intrinsic-independent space, i.e., the normalized plane of the camera pinhole model. 
The point cloud is projected as:
\begin{equation}
    {\left[{\begin{array}{*{20}{c}}
		{\overline{x_i}}, {\overline{y_i}}, {1} 
\end{array}} \right]}^T= \frac{1}{z_i} {\left[{\begin{array}{*{20}{c}}
		{{x}_i}, {{y}_i}, {z_i} 
\end{array}} \right]}^T,
\end{equation}
where $p_i={(x_i,y_i,z_i)}^T$ are the 3D coordinates of the $i$-th point, and $\overline{{p}_{i}}={(\overline{x_i},\overline{y_i},1)}^T$ are the coordinates on the normalized plane. 
In addition, the image is inverse-projected as:
\begin{equation}
    {\left[{\begin{array}{*{20}{c}}
		{\overline{u}}, {\overline{v}}, {1} 
\end{array}}\right]}^T = {K_c}^{-1}{\left[{\begin{array}{*{20}{c}}
		{{u}}, {{v}}, {1} 
\end{array}} \right]}^T,
\end{equation}
where $K_c$ is the intrinsic matrix of the camera, while ${(u,v)}^T$ is the 2D coordinates ${{o}_{i}}$ of the $i$-th pixel on the pixel plane. ${(\overline{u},\overline{v},1)}^T$ are the coordinates on the normalized plane and represented by $\overline{{o}_{i}}$.
\begin{figure}[t]
	\centering
	\vspace{0mm}
	\resizebox{1.0\linewidth}{!}
	{
        \includegraphics[scale=1.0]{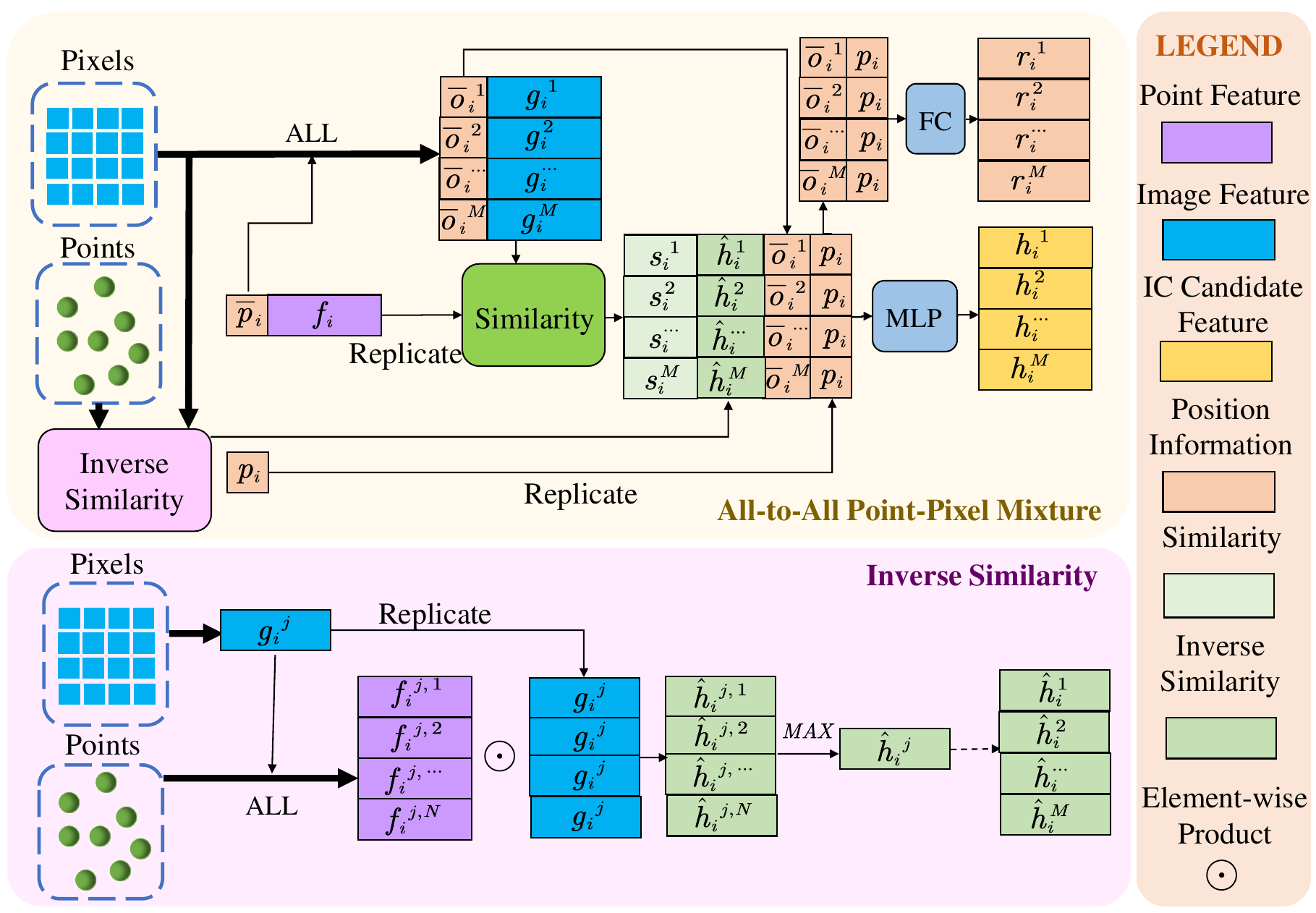}}
	\vspace{-6mm}
	\caption{
		The all-to-all point-pixel mixture for the coarse registration. The all-to-all mixture treats all the pixels as the correspondence candidates of the point. In addition, the inverse similarity is adopted for robust correspondence generation.}
	\vspace{-2mm}
	\label{fig:cv2}
\end{figure}
After the projection and inverse projection, both points and pixels are on the normalized plane. Their coordinates $\overline{{o}_{i}}$ and $\overline{{p}_{i}}$ are independent of the intrinsic.
\modifynew{Then, Implicit Correspondence (IC) generation module calculates the feature similarities of the point-pixel pairs and generates the IC features of each point according to the similarities. Notably, we generate the point-wise IC features rather than pixel-wise IC features, since the 6-DoF relative pose should be regressed from the features of points in 3D space rather than pixels on 2D plane.}

\modifynew{As shown in Fig.~\ref{fig:cvow}, two different types of point-pixel mixtures are proposed to generate the point-pixel pairs. The first is the all-to-all point-pixel mixture, which selects all the pixels to form the point-pixel pairs. The second is the $K$ Nearest Neighbors (KNN)-based point-pixel mixture, which selects the nearest pixel neighbors towards each point on the normalized plane of the pinhole camera model to form the point-pixel pairs.} \modifynew{The appropriate type of point-pixel mixture is chosen according to the initial misalignment between the image and the point cloud. For the task whose initial misalignment is large, the KNN-based mixture can not find the correspondence. Therefore, the all-to-all mixture is adopted for the coarse registration, while the KNN-based mixture is adopted for the fine registration. For the task whose initial misalignment is small, the KNN-based mixture is adopted in both the coarse and fine registrations for the finer correspondence generation.}

\begin{figure}[t]
	\centering
	\vspace{0mm}
	\resizebox{1.0\linewidth}{!}
	{
\includegraphics[scale=1.0]{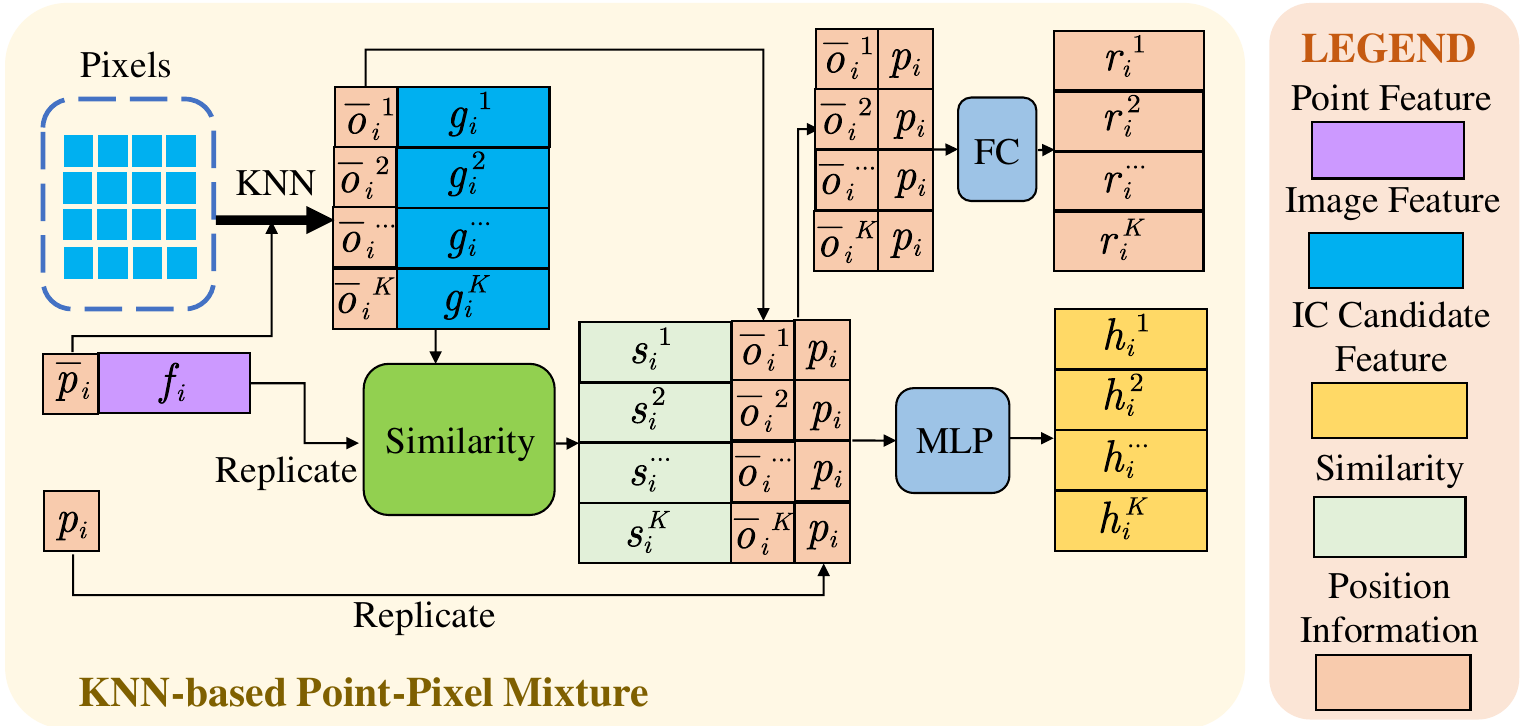}}
	\vspace{-6mm}
	\caption{The KNN-based point-pixel mixture. This type of point-pixel mixture treats the $K$ nearest pixel neighbors as the pixel correspondence candidates. }
	\vspace{-2mm}
	\label{fig:cv1}
\end{figure}

For the all-to-all point-pixel mixture, as shown in Fig.~\ref{fig:cv2}, all the pixels are selected as the matching candidates for the $i$-th point. The position information and image features of the pixel candidates are $\{\overline{{o}}_{i}^{k}\}_{k=1}^M$ and $\{{g}_{i}^{k}\}_{k=1}^{M}$ respectively.
For the KNN-based point-pixel mixture, as shown in Fig.~\ref{fig:cv1}, $K$ nearest pixel neighbors on the normalized plane are selected for each point $\overline{{p}_{i}}$ through KNN. $\{\overline{{o}}_{i}^{k}\}_{k=1}^K$ and $\{{g}_{i}^{k}\}_{k=1}^{K}$ represent the position information and image features of the $K$ nearest pixels respectively. Then, the similarity between ${f}_{i}$ and ${g}_{i}^{k}$ are calculated for the implicit correspondence generation. The similarity is calculated as the element-wise product of the normalized feature vectors. The formula is:
\begin{equation}
       {s}_{i}^{k}=\frac{{f}_{i} - \mu\left(f_i\right)}{\sigma\left(f_i\right)}\odot\frac{{g}_{i}^k - \mu\left(g_i^k\right)}{\sigma\left(g_i^k\right)},
\end{equation}
where ${s}_{i}^{k}$ is the similarity between the features of the $k$-th pixel candidate and its center point. $\odot$ is the element-wise production. $\mu\left(f_i\right)$ and $\sigma\left(f_i\right)$ are the mean and standard deviation of the point feature vector $f_i$, while $\mu\left(g_i^k\right)$ and $\sigma\left(g_i^k\right)$ are the mean and standard deviation of the image feature vector $g_i^k$. In addition, for the all-to-all mixture, the  single-direction similarity from point to pixel can be influenced by the spatial similarity of image features, which can result in the wrong implicit correspondences. Therefore, inspired by \cite{wang2022matters}, we utilize the inverse similarity for the all-to-all pattern to make the IC generation more reliable. The inverse similarity is the maximal pixel-to-point similarity among all the pixels. Specifically, for the $i$-th point, the $j$-th pixel candidate selects all the points as the inverse matching candidates. The point features of the inverse candidates are $\{f_{i}^{j,l}\}_{l=1}^N$. Then, the inverse similarity $\hat{h}_{i}^j$ of the $j$-th pixel candidate is calculated by:
\begin{equation}
\hat{h}_{i}^j = MaxPool(\{ f_i^{j,l} \odot g_i^j\}_{l=1}^N).
\end{equation}
Based on the inverse similarity, only the point-pixel pair that has high similarities in both the forward and inverse directions can be as the correct correspondence.

 The IC candidate features $\{{h}_{i}^{k}\}_{k=1}^{K_1}$ between the $i$-th point and its $k$-th pixel candidate are generated as the Eq. (\ref{eq:c1}) for the all-to-all pattern and Eq. (\ref{eq:c2}) for the KNN pattern respectively. $K_1$ is the number of pixel candidates of each point, which is $M$ for the all-to-all mixture or $K$ for the KNN-based mixture. The equations of IC candidate feature generation are: 
\begin{align}
{h}_{i}^{k}&=MLP({s}_{i}^{k}\oplus\hat{h}_{i}^{k} \oplus \overline{{o}}_{i}^{k} \oplus {p}_{i}), \label{eq:c1} \\{h}_{i}^{k}&=MLP({s}_{i}^{k}\oplus \overline{{o}}_{i}^{k} \oplus {p}_{i}  ),\label{eq:c2}
\end{align}
where $\oplus$ represents the concatenating operation. \modifynew{Notably, the image and point positions, $\overline{{o}}_{i}^{k}$ and ${p}_{i}$, are embedded in the IC features for spatial transformation estimation in the subsequent modules.}

The next step is to estimate the mask ${w}_{i}^{k}$, which is the correspondence salience of the $k$-th IC candidate feature for the $i$-th point. The formula is:
\begin{equation}
{w}_{i}^{k}=Softmax(MLP({h}_{i}^{k} \oplus r_i^k)),
\end{equation}
where $r_i^k=FC({p}_{i} \oplus \overline{o}_{i}^{k})$ is the position embedding, and $FC$ is the fully connected layer to further encode the position information. In addition, $Softmax$ is adopted to normalize the saliences. Then, the weighted sum of IC candidate features with the correspondence salience is performed to generate the IC feature ${ic}_{i}$ for the $i$-th point, as follows:
\begin{equation}
    {ic}_{i}=\sum\limits_{k=1}^{K_1}({h}_{i}^{k}\odot{w}_{i}^{k}).
\end{equation}
The weighted sum makes each point softly match the pixel candidates and resolves the challenge of precise explicit correspondence estimation.

\modifynew{In the following PST embedding module, the point-wise IC features are further gathered to embed the spatial transformation of the point-pixel patch in PST embedding features for the outlier mask prediction.} 
Specifically, the projection-aware grouping-based neighborhood query is adopted to query each point's $K_2$ nearest point neighbors.
The position information $\{p_i^m\}_{m=1}^{K_2}$ and the IC features $\{ic_i^m\}_{m=1}^{K_2}$ of the point neighbors are gathered. 
In addition, inspired by RandLA-Net~\cite{hu2020randla}, the relative position information between each point neighbor and its center point is calculated to enrich the position information. 
The overall position information $\{{u}_{i}^{m}\}_{m=1}^{K_2}$ are composed of four parts: The absolute coordinates of the center point, the absolute coordinates of point neighbors, the relative coordinates, and the Euclidean distance, as follows:
\begin{equation}
    {u}_{i}^{m}={p}_{i}\oplus{p}_{i}^{m}\oplus\left({p}_{i}^{m}-{p}_{i}\right)\oplus\lVert{p}_{i}^{m}-{p}_{i}\rVert,
\end{equation}
where $\lVert\cdot\rVert$ represents the $L2$-norm. The overall position information is fed into an FC layer to obtain the position embedding ${b}_{i}^{m}$ as follows:
\begin{equation}
    {b}_{i}^{m}=FC({u}_{i}^{m}).
\end{equation}
Then, the weighted \modify{PST estimation} is performed. The point feature ${f}_{i}$ of the $i$-th center point, the position embedding ${b}_{i}^{m}$ of the $m$-th point neighbor, and the IC feature ${ic}_{i}^{m}$ of the $m$-th point neighbor are concatenated and fed into a shared MLP block to calculate the weights:
\begin{equation}
    {w}_{i}^{m}=Softmax(MLP({f}_{i}\oplus{b}_{i}^{m} \oplus{ic}_{i}^{m})),
\end{equation}
where ${w}_{i}^{m}$ is the weight of the the $m$-th point neighbor's IC feature for the $i$-th center point. Based on the estimated weights, the weighted sum of the IC features is performed to estimate the \modify{patch spatial transformation}, as follows:
\begin{equation}
    {e}_{i}=\sum\limits_{m=1}^{K_2}({ic}_{i}^{m}\odot{w}_{i}^{m}),
\end{equation}
where ${e}_{i}$ is the \modify{PST embedding features} of the $i$-th point. Therefore, $N$ \modify{PST embedding features} are obtained as the 2D-3D cost volumes, which are represented by $E=\{{e}_{i}|i=1,2,...,N\}$. Specifically, \modifylatest{${E}^{c}$} represents the 2D-3D cost volumes estimated in the coarse registration, and \modifylatest{${E}^{f}$} represents the 2D-3D cost volumes estimated in the fine registration.
\subsection{Coarse Registration}\label{section:coarse}
\subsubsection{Context Gathering Module} 
To localize the robot within a large range, the context gathering module further gathers the 2D-3D cost volumes for the coarse registration. The 2D-3D association information in the context is utilized to refine the 2D-3D cost volume of each point.

Specifically, we use the same feature extraction module in the point cloud feature extraction to gather the 2D-3D cost volumes \modifylatest{$E^c$}. The gathered 2D-3D cost volumes are \modifylatest{${E}^{c}_{new} =\{{e}_{new,i}^{c}|i=1,2...,{N^4}\}$}, which is calculated as:
\begin{equation}
	\modifylatest{e^c_{new,i}} = MaxPool(MLP(\{\modifylatest{e^c_{i,k}}\}^{K^4}_{k=1})),
\end{equation}
where the center points of each point group are sampled using the same indexes as the fourth layer of the point cloud feature extraction.

\subsubsection{Outlier Mask Prediction Module}\label{section:mp}
The 2D-3D cost volumes \modifylatest{$E^c_{new}$} embed the information of the 2D-3D association. However, the outliers of the 2D-3D association, such as  
the points out of the image frustum~\cite{li2021deepi2p} which have no pixel correspondence, should be filtered.
The former works~\cite{ren2022corri2p,feng20192d3d} utilize the RANSAC to filter the outliers for relative pose estimation. In this paper, 
the outlier masks are learned to filter the outliers based on the point features and PST embedding features.

In detail, in the coarse registration, the outlier mask prediction module uses the point features of the fourth layer $F^4=\{{f}_{i}^{4}|i=1,2...,{N^4}\}$ and context-gathered 2D-3D cost volumes \modifylatest{${E}^{c}_{new}$} to estimate the coarse outlier masks \modifylatest{$M^c=\{{m}_{i}^{c}|i=1,2...,{N^4}\}$}. The formula is: 
\modifylatest{\begin{equation}
m_i^c=MLP({e}^{c}_{new,i}\oplus f_i^4).
\end{equation}}

\subsubsection{Pose Regression}\label{section:pr}
The pose regression first aggregates the 2D-3D cost volumes weighted by the outlier masks to generate the \modify{spatial transformation embedding feature}.  Specifically, the weights \modifylatest{$MW^c=\{{mw}_i^c\}_{i=1}^{N^4}$} are calculated by performing softmax on the outlier masks \modifylatest{$M^c$}. The outlier masks assign outliers with low weights. Thus, the inlier 2D-3D cost volumes are mainly aggregated. Then, the FC layers decode the spatial transformation embedding feature and obtain the relative pose between the image and point cloud.  In detail, an FC layer is first adopted to perform the linear transformation on the spatial transformation embedding feature to gain the middle feature. Then, the dropout operation is performed on the middle feature during the training to overcome overfitting. Finally, the predicted quaternion $q$ and the translation vector $t$ are obtained by the linear transformation of two FC layers respectively on the middle feature. The predicted quaternion and translation vector in the coarse registration are denoted as \modifylatest{$q_c$ and $t_c$}. The formula of the pose regression is:
\modifylatest{\begin{equation}
q_c = \frac{FC(\sum\limits_{i=1}^{N^4}({e}_{new,i}^{c}\odot{mw}_{i}^{c}))}{\lvert FC(\sum\limits_{i=1}^{N^4}({e}_{new,i}^{c}\odot{mw}_{i}^{c}))\rvert},
\end{equation}
\vspace{-2mm}
\begin{equation}
t_{c} = FC(\sum\limits_{i=1}^{N^4}({e}_{new,i}^{c}\odot{mw}_{i}^{c})),
\end{equation}}
where the quaternion is normalized to gain a unique representation of the rotation. \modifylatest{The coarse relative pose $q_c,t_c$ is treated as the initial pose estimation $q_{f}^0,t_{f}^0$ of the fine registration.}

 \subsubsection{Upsampling Layer}
Two upsampling layers are adopted to upsample the coarse 2D-3D cost volumes $E_{new}^c$ and outlier masks $M^c$ respectively. The upsampling converts the prior knowledge of coarse registration. In the upsampling layers, the projection-aware neighborhood query is adopted to group $K$ nearest points $\{p_{i,k}^4\}_{k=1}^K$ from the point cloud $P^4$ for each point $p_i^3$ in the point cloud $P^3$. 
After querying, the relative position between each grouped point and its center point is calculated. 
Then, the coarse features of grouped points are concatenated with the relative position. 
The concatenated features are aggregated by a shared MLP block and a max-pooling operation. After the aggregation, the aggregated features and the point features $F^3$ are concatenated and fed into an FC layer. Finally, the upsampled 2D-3D cost volumes \modifylatest{$UE=\{{ue}_{i}|i=1,2,...,{N^3}\}$} and outlier masks \modifylatest{$UM=\{{um}_{i}|i=1,2,...,{N^3}\}$} are outputted by the two upsampling layers respectively. The formulas are:
\begin{align}
        \modifylatest{ue_i} = FC(f_i^3&\oplus MaxPool(MLP(\{\modifylatest{{e}_{new,i,k}^c}\notag\\&\oplus (p_{i,k}^4-p_i^3)\}_{k=1}^{K}))),\\
    \modifylatest{um_i} = FC(f_i^3&\oplus MaxPool(MLP(\{\modifylatest{{m}_{i,k}^c}\notag\\&\oplus (p_{i,k}^4-p_i^3)\}_{k=1}^{K}))).
\end{align}
\modifylatest{$UE$ and $UM$} will be used in the cost volume optimization and outlier mask prediction in the fine registration.

\subsection{Fine Registration}\label{section:finer}

\subsubsection{Pose Warping}
To gain an image-point cloud pair with smaller misalignment based on the coarse prior knowledge, we utilize the relative pose predicted in the coarse registration to warp the point cloud $P^3$. The formula is as follows:
\begin{equation}
(0,{P}^{3}_{warped})=\modifylatest{{q}_{f}^{i}}(0,{P}^{3})(\modifylatest{{q}_{f}^{i}})^{-1}+(0,\modifylatest{{t}_{f}^{i}}),
\end{equation}
where ${P}^{3}_{warped}$ are the warped 3D coordinates of $P^3$ and \modifylatest{${q}_{f}^{i},t_{f}^{i}$} are the estimated relative pose \modifylatest{in the last iteration, where $i\in\{0,1,\cdots,K_{iter}-1\}$}. $P_{warped}^3$, $F^3$, $O^3$, and $G^3$ are fed into the 2D-3D cost volume module using KNN-based point-pixel mixture to obtain the fine 2D-3D cost volumes \modifylatest{$E^f$}.
\subsubsection{Optimization Module}
The fine 2D-3D cost volumes \modifylatest{$E^f$} are optimized with upsampled 2D-3D cost volumes \modifylatest{$UE$} to gain the optimized 2D-3D cost volumes \modifylatest{$OE=\{{oe}_{i}|i=1,2,...,{N^3}\}$}. The optimization enables the fine registration to utilize coarse prior knowledge. Specifically,
$E^f$, $UE$, and the point features of the third layer $F^3$ are fed into a shared MLP block to gain $OE$. The formula is:
\begin{equation}
\modifylatest{oe_i}=MLP(\modifylatest{e^f_i}\oplus \modifylatest{ue_i} \oplus f^3_i).
\end{equation}
Then, the outlier masks of the fine registration \modifylatest{$M^f=\{{m}^f_{i}|i=1,2,...,{N^3}\}$} are predicted by the outlier mask prediction module. In addition, the upsampled coarse outlier masks \modifylatest{$UM$} are utilized to optimize the outlier mask prediction by coarse prior knowledge of outlier estimation. Specifically, \modifylatest{$OE$, $UM$}, and the point features of the third layer $F^3$ are fed into a shared MLP block to obtain \modifylatest{$M^f$}, as follows:
\begin{equation}
\modifylatest{m^f_i}=MLP(\modifylatest{oe_i}\oplus \modifylatest{{um}_i} \oplus f^3_i).
\end{equation}
\subsubsection{Pose Refinement}
The pose regression proposed in Section \ref{section:pr} is adopted to regress the residual relative pose \modifylatest{$\Delta{q}$ and $\Delta{t}$} in the fine registration. The formula is:
\modifylatest{\begin{equation}
\Delta q = \frac{FC(\sum\limits_{i=1}^{N^3}({oe}_{i}\odot{mw}_{i}^{f}))}{\lvert FC(\sum\limits_{i=1}^{N^3}({oe}_{i}\odot{mw}^f_{i}))\rvert},
\end{equation}
\vspace{-2mm}
\begin{equation}
\Delta t = FC(\sum\limits_{i=1}^{N^3}({oe}_{i}\odot{mw}_{i}^{f})),
\end{equation}}
where the weights \modifylatest{$MW^f=\{{mw}_{i}^{f}\}_{i=1}^{N^3}$} are as well calculated by performing softmax on the outlier masks \modifylatest{$M^f$}. 

The residual relative pose \modifylatest{$\Delta{q}$ and $\Delta{t}$} are predicted after the pose warping on $P^3$ with \modifylatest{${q}_{f}^i$ and ${t}_{f}^i$}. Therefore,  \modifylatest{$\Delta{q}$ and $\Delta{t}$} refine \modifylatest{${q}_{f}^{i}$ and ${t}_{f}^{i}$} respectively to obtain the refined relative pose \modifylatest{${q}_{f}^{i+1}$ and ${t}_{f}^{i+1}$ in the $i$-th iteration}, as follows:
\modifylatest{\begin{equation}
{q}_{f}^{i+1}=\Delta{q}\cdot {q}_{f}^{i},
(0,{t}_{f}^{i+1})=\Delta{q}(0,{t}_{f}^{i})\Delta{q}^{-1}+(0,\Delta{t}).
\end{equation}}
\subsection{Loss Function}\label{section:loss}
The training loss is calculated after the forward path by comparing the output of the network and the ground truth. 
Inspired by LO-Net~\cite{li2019net}, two learnable parameters $s_q$ and $s_t$ are adopted to the loss function to bridge the scale and unit difference between the quaternion and translation vector. The loss function of single registration is as follows:
\begin{equation}
\begin{aligned}
   \mathcal{L}({q},{t},{q}_{gt},{t}_{gt})&={\lVert{q}_{gt}-{q}\rVert}_{2}\cdot {e}^{-{s}_{q}}+{s}_{q}\\&+{\lVert t_{gt}-{t}\rVert}_{1}\cdot{e}^{-{s}_{t}}+{s}_{t}, 
\end{aligned}
\end{equation}
where ${q}_{gt}$ is the ground truth quaternion, and ${t}_{gt}$ is the ground truth translation vector. ${\lVert\cdot\rVert}_1$ and ${\lVert\cdot\rVert}_2$ represent the $L1$-norm and $L2$-norm respectively.

During the training, the number of iteration is set as one to boost the training efficiency. The total registration loss is composed of both losses of the coarse and fine registrations, as follows:
\modifylatest{\begin{equation}
\mathcal{L}={\alpha_c}\cdot\mathcal{L}({q}_{c},{t}_{c},{q}_{gt},{t}_{gt})+{\alpha_f}\cdot\mathcal{L}({q}_{f},{t}_{f},{q}_{gt},{t}_{gt}),
\end{equation}}
where $\mathcal{L}$ is the total registration loss of the network, and \modifylatest{${\alpha_c}$ and ${\alpha_f}$} are the weights of the single registration losses for the coarse registration and fine registration respectively.

\section{Experiments}\label{section:exp}
In this section, we conduct the experiments to answer the four questions and evaluate the effectiveness and efficiency of our end-to-end 2D-3D registration architecture I2PNet:
\begin{itemize}
	\item Can end-to-end 2D-3D registration architecture localize the robot within a large range?
	\item Can end-to-end 2D-3D registration architecture outperform the end-to-end 2D-2D registration architectures?
	\item How does each module in I2PNet contribute to the end-to-end 2D-3D registration for robot localization?
	\item \modifylatest{Can end-to-end 2D-3D registration architecture perform efficient image-based robot localization?}
\end{itemize}
\subsection{Implementation Details}\label{implement details}
We conduct all the experiments on an NVIDIA GeForce RTX 3090. PyTorch~\cite{paszke2019pytorch} is adopted to develop I2PNet. The batch sizes for both training and testing are set as $8$. The Adam~\cite{kingma2014adam} optimizer with ${\beta}_{1}=0.9$ and ${\beta}_{2}=0.999$ is adopted in the training. In addition, the initial learning rate of the optimizer is set as $10^{-3}$, and the learning rate delays by $1\%$ after each epoch. 
The weights in the total loss function are ${\alpha_3}=0.8,\ {\alpha_4}=1.6$. The learnable parameters ${s}_{q}$ and ${s}_{t}$ are initialized as $-2.5$ and $0$ respectively. Moreover, the dropout rate for the pose regression is set as $0.5$. \modifylatest{The hyperparameters of the modules in I2PNet are listed in the appendix.}
In the feature extraction of the LiDAR point cloud, the point coordinates are maintained in the LiDAR coordinate system, ensuring that the features do not change with pose transformations of the LiDAR point cloud. 
\modifylatest{As one iteration in fine registration already achieves state-of-the-art performance with the highest efficiency compared to previous methods, we set the number of fine registration iterations $K_{iter}$ as one when comparing with other methods. In Sec.~\ref{sec:itera_fine}, we will show detailed experiments analyzing the trade-off between accuracy and efficiency.}

In the experiments, we divide the vehicle localization tasks into two parts, including large-range localization and small-range localization, since the experiment conditions and methods for comparison differ in the two parts. For large-range localization, existing methods adopt 2D-3D registration architecture~\cite{elbaz20173d,yew20183dfeat,li2019usip,bai2020d3feat,huang2021predator,li2021deepi2p,ren2022corri2p} and preserve the complete 3D point cloud for the image-to-point cloud association. This enables the large-range localization methods to localize within large errors. Thus, the localization range for large-range localization is within $360^\circ$ and $10m$.
As for small-range localization, existing methods adopt 2D-2D coarse-to-fine registration architecture~\cite{cattaneo2019cmrnet,chang2021hypermap,chen2022i2d} and utilize the LiDAR depth image as the network input. These methods rely on the co-visible region between the LiDAR depth image and the actual RGB image under small range of errors. Thus, the localization range for small-range localization is within the random rotation inside $[-10^\circ,10^\circ]$ and random translation inside $[-2m,2m]$.

\setlength{\tabcolsep}{3mm}
\begin{table*}[t]
    \centering
    \caption{Large-Range Localization Error on the KITTI Odometry and NuScenes Datasets}
    \resizebox{0.7\textwidth}{!}
	{
    \begin{tabular}{c||cc|cc}
\toprule
\multirow{2}{*}{\begin{tabular}{c}Method \end{tabular}} & \multicolumn{2}{c|}{KITTI Odometry} & \multicolumn{2}{c}{nuScenes} \\
& RRE ($^\circ$)~$\downarrow$ & RTE (m)~$\downarrow$ & RRE ($^\circ$)~$\downarrow$ & RTE (m)~$\downarrow$ \\
\hline
\hline
\noalign{\smallskip}
Grid. Cls. + EPnP~\cite{li2021deepi2p} & 6.48 $\pm$ 1.66 & 1.07 $\pm$ 0.61 & 7.20 $\pm$ 1.65 & 2.35 $\pm$ 1.12 \\

DeepI2P (3D)~\cite{li2021deepi2p} & 6.26 $\pm$ 2.29 & 1.27 $\pm$ 0.80 & 7.18 $\pm$ 1.92 & 2.00 $\pm$ 1.08 \\
DeepI2P (2D)~\cite{li2021deepi2p} & 4.27 $\pm$ 2.29 & 1.46 $\pm$ 0.96 & 3.54 $\pm$ 2.51 & 2.19 $\pm$ 1.16 \\
CorrI2P~\cite{ren2022corri2p} & 2.07 $\pm$ 1.64 & 0.74 $\pm$ 0.65 & 2.65 $\pm$ 1.93 & 1.83 $\pm$ 1.06 \\
\midrule
Ours (I2PNet) & \textbf{0.83 $\pm$ 1.04} &
\textbf{0.21 $\pm$ 0.29} &
\textbf{1.13 $\pm$ 1.08}&
\textbf{0.75 $\pm$ 0.59}\\
\bottomrule
\end{tabular}
}
    \label{tab:general_results}
    \vspace{-3mm}
\end{table*}
\begin{figure*}[h]
    \centering
	\includegraphics[width=1.0\textwidth]{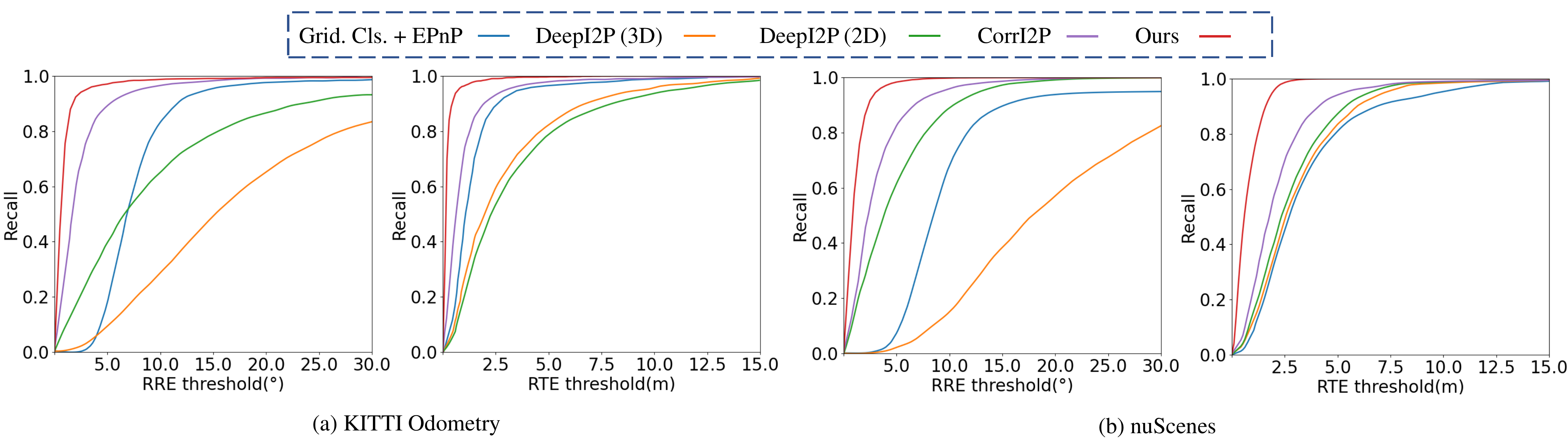}
    \vspace{-6mm}
    \caption{The registration recall curves of the methods with different RTE and RRE thresholds on KITTI Odometry and nuScenes datasets. The y-axis (recall) of the recall curve presents the success rate that RREs or RTEs are less than the threshold in the x-axis.}
	\vspace{-3mm}
    \label{fig:recall}
\end{figure*}

\begin{figure}[h]
    \centering
	\includegraphics[width=0.8\linewidth]{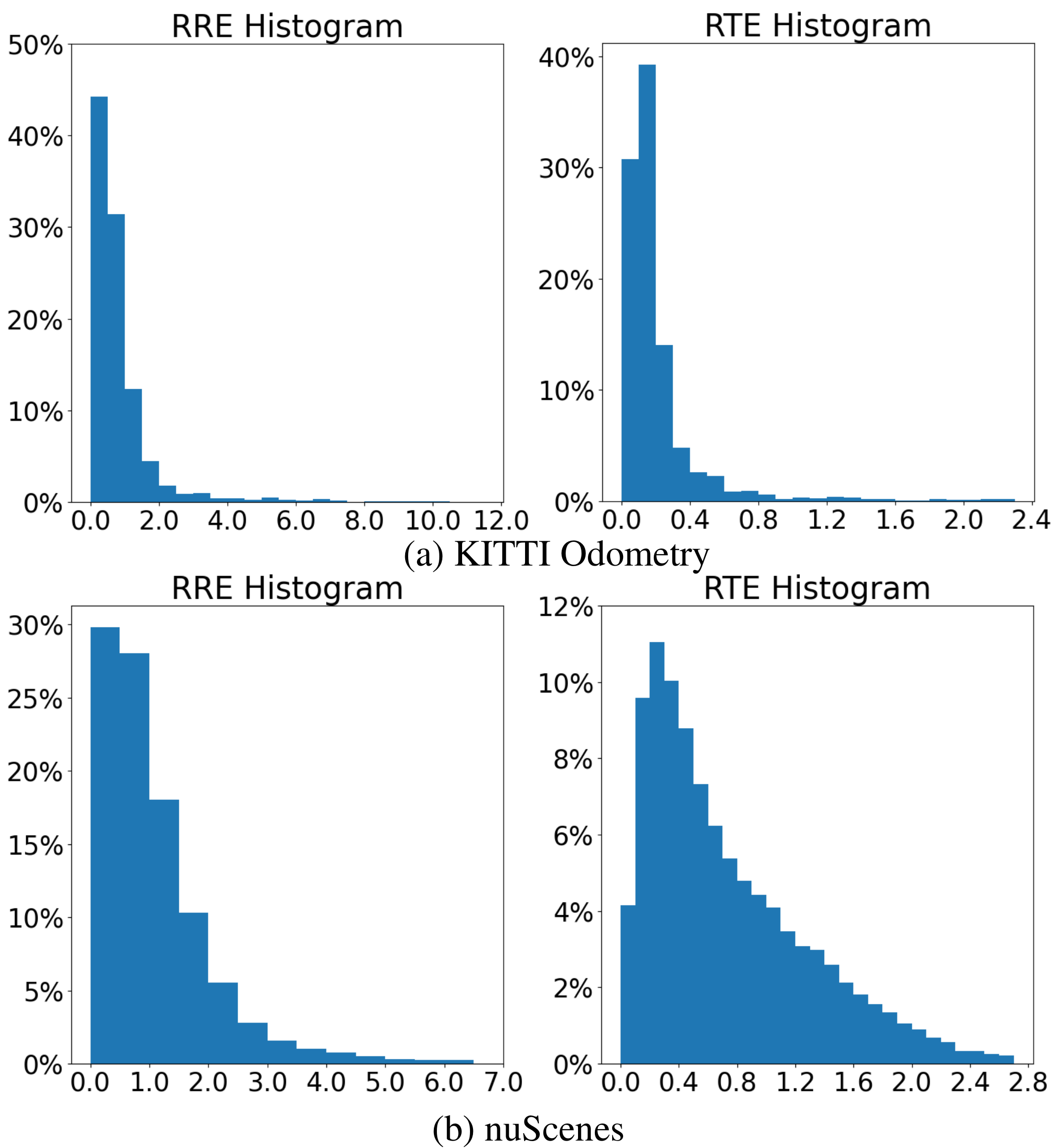}
    \caption{The histograms of RTE and RRE on the KITTI Odometry and nuScenes datasets. The x-axis is the RTE (m) or RRE ($^\circ$), and the y-axis is the percentage falling into the corresponding bin. The bin size of RRE and RTE is $0.5^\circ$ and $0.1m$ respectively. 
	}
    \label{fig:hist}
	\vspace{-3mm}
\end{figure}

\begin{figure*}[t]
\centering
\vspace{0mm}

\resizebox{1.0\textwidth}{!}
{
	\includegraphics[scale=1.0]{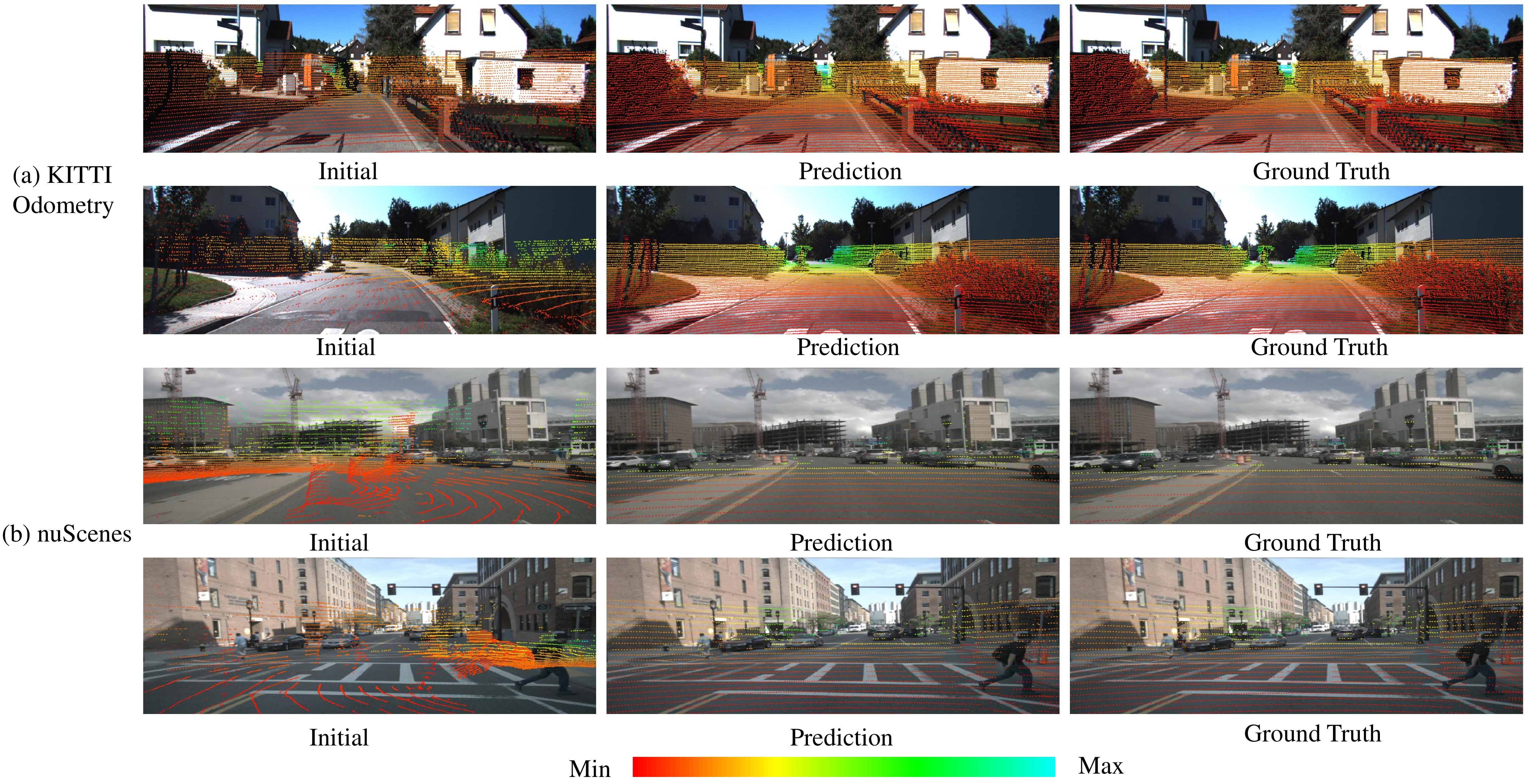}}
\vspace{-3mm}
\caption{Visualization of large-range localization results on KITTI Odometry \modifynew{and nuScenes} datasets. This figure presents the large-range localization results on the KITTI Odometry \modifynew{and nuScenes} datasets through the visualization of image-to-point cloud registration. The color bar shows the depth of each LiDAR point.
}
\vspace{-2mm}
\label{fig:kd_vis}
\end{figure*}

\begin{figure*}[t]
\centering
\vspace{0mm}
\resizebox{1.0\textwidth}{!}
{
	\includegraphics[scale=1.0]{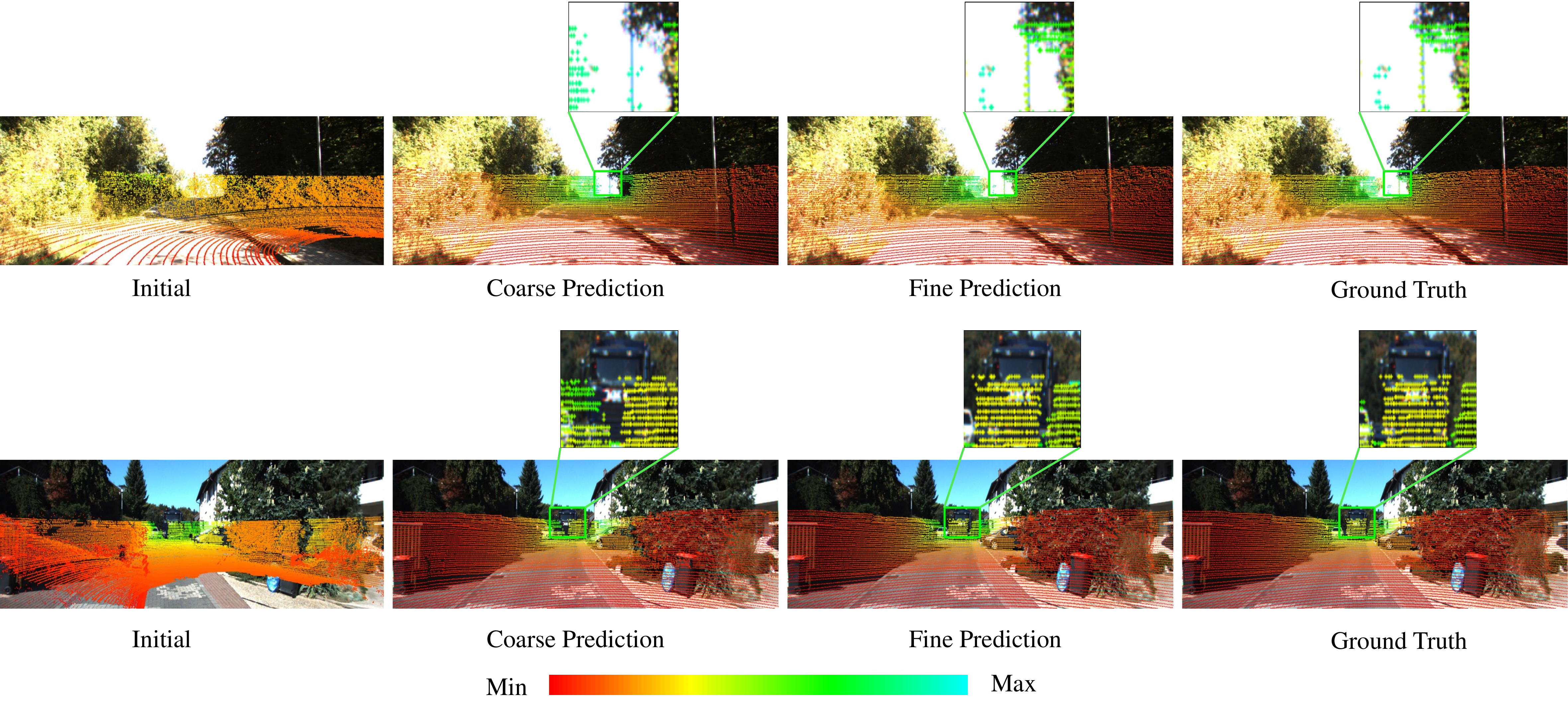}}
\vspace{-3mm}
\caption{Visualization of coarse and fine registration results. The coarse and fine predictions are obtained from the coarse and fine registration respectively. The zoom-in views of the areas marked by the bounding boxes in each picture are to better present the difference among the coarse prediction, fine prediction, and ground truth. The color bar has the same meaning as Fig. \ref{fig:kd_vis}.
}
\vspace{-2mm}
\label{fig:kd_vis_cf}
\end{figure*}

\subsection{Large-Range Localization}
\subsubsection{Dataset and Data Pre-processing}
The experiments are performed on the KITTI Odometry dataset~\cite{geiger2013vision} and nuScenes dataset~\cite{nuscenes2019}.
In the KITTI Odometry dataset, the training set contains 0-8 sequences, and the test set includes 9-10 sequences. In the nuScenes dataset, we refer to the official split to use the 850 traversals to train our model, and 150 traversals are left for testing.
The following data preprocessing are performed for the large-range localization task:

\textbf{KITTI Odometry dataset.} We select the image-point cloud pairs from the same frame. In this setting, the image and point cloud are captured simultaneously by the RGB camera and the LiDAR that have a fixed relative position. For the LiDAR point cloud map generation, a random transformation is generated as the pose of the camera in the map coordinate system. The point cloud is transformed by the generated transformation to the map coordinate system. \modify{To localize robots with various orientations and a large range of displacement, the random transformation contains a rotation around the up-axis within $[-\pi,\pi]$ and a 2D translation on the ground within the range of $10\ m$.} The network is expected to predict the relative pose between the LiDAR point cloud map and the image to localize the robot. In addition, the top 50 rows of each image are cropped because they are occupied by the sky without corresponding LiDAR points. After the cropping, the image is resized to $160\times512$. As for the 
	point cloud, we input all the points in the LiDAR point cloud map. For the calculation of the 2D spherical coordinates, the initial upper bounds $(H,W)$ are $(64,1800)$. The up and down vertical field-of-views are $f_{up}=2.0,f_{down}=24.8$.
    Moreover, for a fair comparison with other methods, each feature vector of the initial point features $F^0$ is the concatenation of the estimated surface normal vector and the intensity.
 
    \textbf{NuScenes dataset.} 
    The image and point cloud in the nearby frame are selected to form the image-point cloud pair. The known relative pose between the point cloud and the image is used to transform the point cloud. Thus, the aligned image-point cloud pair is obtained. 
    We use the same method to form the LiDAR point cloud map as the KITTI Odometry dataset. The ranges of the rotation and translation are as well the same. In addition, the top 100 rows of each image are cropped. After the cropping, the image is resized to $160\times 640$. As for the point cloud, all the points in the LiDAR point cloud map are inputed. For the calculation of the 2D spherical coordinates, the initial upper bounds $(H,W)$ are $(32,1800)$. Up and down vertical field-of-views are $f_{up}=10.0,f_{down}=30.0$. Moreover, for a fair comparison with the other methods, each feature vector of the initial point features $F^0$ is the concatenation of a three-dimensional zero vector and the intensity.

\subsubsection{Experimental Results and Visualization}\label{section:general_result}
\modify{To ensure a fair comparison with previous works, we refer to the previous works~\cite{elbaz20173d,yew20183dfeat,li2019usip,bai2020d3feat,huang2021predator,li2021deepi2p,ren2022corri2p} to adopt the Relative Rotational Error (RRE) and the Relative Translation Error (RTE)~\cite{ma2016fast} as the metrics to evaluate the performance of the models, which are calculated as:
\modifynew{\begin{equation}
    RRE = \sum\limits_{i=1}^3\lvert\theta_i\rvert, RTE = \lVert t_{pred}-t_{gt}\rVert_2,
\end{equation}}
where $\{\theta_i\}_{i=1}^3$ are the Euler angles of the rotation error matrix $R_{pred}^{-1}R_{gt}$, where predicted rotation matrix $R_{pred}$ is calculated by the predicted quaternion. $R_{gt}$ is the ground truth rotation matrix of the robot pose. $t_{pred}$ is the predicted translation vector. $t_{gt}$ is the ground truth translation vector of the robot pose.} As CorrI2P~\cite{ren2022corri2p}, we calculate the average RRE and RTE of the samples whose RREs are less than $10^\circ$ and whose RTEs are less than $5m$, and the quantitative results are presented in Table~\ref{tab:general_results}.

In Table~\ref{tab:general_results}, we compare the performance of I2PNet with CorrI2P~\cite{ren2022corri2p} and the three methods proposed in DeepI2P~\cite{li2021deepi2p}. In the three methods in DeepI2P, \emph{Grid. Cls. + EPnP} divides the image to $32\times 32$ patches. Then, the points falling into which patch are predicted. The predictions are used to construct the point-to-pixel correspondences. Finally, the RANSAC-based EPnP is adopted to predict the relative pose. In addition, \emph{DeepI2P (3D)} and \emph{DeepI2P (2D)} predict the points falling into the image frustum, and obtain the optimal relative pose by solving the inverse camera projection problem. Their difference is that \emph{DeepI2P (3D)} sets the relative pose of six Degrees of Freedom (6-DoF) in the optimization solver while \emph{DeepI2P (2D)} sets the up-axis rotation and translation on the ground as the relative pose in the optimization solver.
The results show that I2PNet outperforms the CorrI2P and three methods of DeepI2P on both two datasets. 
\modifynew{This validates that our end-to-end 2D-3D registration architecture can effectively optimize the whole registration process based on the proposed differential 2D-3D cost volume module, which bridges feature extraction and pose regression processes. Therefore, the robot is more effectively localized in a large range than the methods with separate modules.}  

The recall curves on the RRE and RTE on the KITTI Odometry dataset and nuScenes dataset are presented in Fig.~\ref{fig:recall}. The results show that I2PNet has much better recall curves than the other works on both two datasets. This further shows the localization performance of I2PNet is better than the other works. In addition, we present the RRE and RTE histograms of the predictions of I2PNet on the two datasets in Fig.~\ref{fig:hist}. The histograms show that the RREs and RTEs of the predictions are mostly within the smallest error range on both two datasets.

\modifynew{Fig.~\ref{fig:kd_vis} 
qualitatively show I2PNet's performance on KITTI Odometry and nuScenes datasets respectively on the large-range robot localization task.} For large-range robot localization, the initial misalignments between the image and point cloud are terrible as in the visualization. Despite the difficult image-point cloud pairs input, the predictions are close to the ground truths, which indicates the high localization precision of I2PNet.
In addition, we qualitatively show the effectiveness of the fine registration in Fig.~\ref{fig:kd_vis_cf}. The visualization results show that the coarse registration already generates the image-point cloud pair with a small misalignment and presents an acceptable localization precision. Moreover, the fine registration further refines the registration. Thus, the final localization error is smaller. 

\subsubsection{Validation on Non-Car-Like Platform} We utilize M2DGR dataset~\cite{yin2021m2dgr} for ground robot platform validation. M2DGR dataset~\cite{yin2021m2dgr}  is a dataset collected in Shanghai Jiao Tong University (SJTU) campus through a non-car-like ground robot in different scenarios, including the outdoor scenes, hall scenes, room scenes, and the scenes of entering and exiting the lift. The LiDAR point clouds are obtained through a Velodyne VLP-32C LiDAR. Experimental results in Table~\ref{table:m2dgr} demonstrate that our method performs well in indoor and outdoor environments, including outdoors, halls, rooms, and the scenes of entering and exiting the lift. These scenes are more diverse and complex compared to autonomous driving scenarios, and our method significantly outperforms other approaches.

\setlength{\tabcolsep}{1mm}
\begin{table*}[t]
		\caption{\modifylatest{Large-Range Localization Evaluated on M2DGR Dataset. All Models Are Trained and Tested on M2DGR Dataset to Demonstrate the Performance on a Non-Car-Like Platform, a Ground Robot Platform}}
	\begin{center}
\resizebox{0.9\linewidth}{!}
		{
\modifylatest{\begin{tabular}{l||cc|cc|cc|cc}
\toprule 
\multirow{-0.5}{*}{\begin{tabular}[c]{@{}c@{}}Method \end{tabular}}
& \multicolumn{2}{c|}{Outdoor} & \multicolumn{2}{c|}{Hall}& \multicolumn{2}{c|}{Lift} & \multicolumn{2}{c}{Room}\\
&RRE~($^\circ$)~$\downarrow$ & RTE~$(m)$~$\downarrow$ & RRE~($^\circ$)~$\downarrow$ & RTE~$(m)$~$\downarrow$ & RRE~($^\circ$)~$\downarrow$ & RTE~$(m)$~$\downarrow$& RRE~($^\circ$)~$\downarrow$ & RTE~$(m)$~$\downarrow$ \\
\hline
\hline
\noalign{\smallskip}
Grid. Cls. + EPnP~\cite{li2021deepi2p} & 6.49 $\pm$ 2.59 & 2.50 $\pm$ 1.22 & 6.71 $\pm$ 2.29 & 2.61 $\pm$ 1.33 & 5.33 $\pm$ 2.90 & 3.12 $\pm$ 1.09 & 4.34 $\pm$ 1.79 & 2.59 $\pm$ 0.75\\
DeepI2P (3D)~\cite{li2021deepi2p} & 5.96 $\pm$ 0.84 & 1.10 $\pm$ 1.01 & 6.92 $\pm$ 2.00 & 1.62 $\pm$ 1.23 & 6.20 $\pm$ 2.90 & 1.51 $\pm$ 0.62 & 5.56 $\pm$ 1.62 & 0.42 $\pm$ 0.09\\
DeepI2P (2D)~\cite{li2021deepi2p} & 5.57 $\pm$ 2.54 & 2.18 $\pm$ 1.16 & 4.70 $\pm$ 2.83 & 2.20 $\pm$ 1.20 & 5.06 $\pm$ 2.70 & 1.82 $\pm$ 0.99 & 5.42 $\pm$ 2.23 & 1.82 $\pm$ 0.52 \\\
CorrI2P~\cite{ren2022corri2p} & 5.12 $\pm$ 2.30 & 2.17 $\pm$ 1.05 & 5.10 $\pm$ 2.40 & 1.53 $\pm$ 0.93 & 5.81 $\pm$ 2.27 & 2.17 $\pm$ 1.05 & 6.98 $\pm$ 1.84 & 1.18 $\pm$ 0.88 \\
\midrule
Ours (I2PNet) & \textbf{1.13 $\pm$ 0.84} &
\textbf{0.29 $\pm$ 0.18} &
\textbf{0.89 $\pm$ 0.80} &
\textbf{0.24 $\pm$ 0.15} &
\textbf{1.51 $\pm$ 1.36} &
\textbf{0.36 $\pm$ 0.24} &
\textbf{1.70 $\pm$ 1.36} &
\textbf{0.31 $\pm$ 0.22} \\
				\bottomrule
			\end{tabular}}
		}
	\end{center}
	\label{table:m2dgr}
\end{table*}

\setlength{\tabcolsep}{1mm}
\begin{table*}[t]
\renewcommand{\arraystretch}{1.1}
		\caption{\modifylatest{Generalization Experiments from KITTI Odometry Dataset to NuScenes and M2DGR Datasets. All Models Are Trained on KITTI Odometry Dataset and Tested on All the Other Datasets}}
	\begin{center}
		\resizebox{1\linewidth}{!}
		{
\modifylatest{\begin{tabular}{l||cc|cc|cc|cc|cc}
\toprule 
\multirow{-0.5}{*}{\begin{tabular}[c]{@{}c@{}}Method \end{tabular}}
& \multicolumn{2}{c|}{nuScenes}& \multicolumn{2}{c|}{M2DGR (Outdoor)} & \multicolumn{2}{c|}{M2DGR (Hall)}& \multicolumn{2}{c|}{M2DGR (Lift)} & \multicolumn{2}{c}{M2DGR (Room)}\\
&RRE~($^\circ$)~$\downarrow$ & RTE~$(m)$~$\downarrow$ & RRE~($^\circ$)~$\downarrow$ & RTE~$(m)$~$\downarrow$ & RRE~($^\circ$)~$\downarrow$ & RTE~$(m)$~$\downarrow$& RRE~($^\circ$)~$\downarrow$ & RTE~$(m)$~$\downarrow$  & RRE~($^\circ$)~$\downarrow$ & RTE~$(m)$~$\downarrow$ \\
\hline
\hline
\noalign{\smallskip}
Grid. Cls. + EPnP~\cite{li2021deepi2p}  & 4.94 $\pm$ 2.84 & 3.22 $\pm$ 1.20 & 6.04 $\pm$ 3.15 & 2.21 $\pm$ 1.23 & 4.46 $\pm$ 2.83 & 3.53 $\pm$ 1.25 & 4.43 $\pm$ 2.65 & 3.43 $\pm$ 1.11 & failed & failed\\
DeepI2P (3D)~\cite{li2021deepi2p}  & failed & failed & failed & failed & failed & failed & failed & failed & failed & failed  \\
DeepI2P (2D)~\cite{li2021deepi2p}  & failed  & \multicolumn{1}{c|}{failed} & failed & failed & failed & failed & failed & failed & failed & failed \\
CorrI2P~\cite{ren2022corri2p} & failed &failed & failed & failed & failed & failed & failed & failed & failed & failed \\
\midrule
Ours (I2PNet) &
\textbf{2.76 $\pm$ 2.13} &
\textbf{2.39 $\pm$ 1.14} &
\textbf{2.71 $\pm$ 2.29} &
\textbf{1.70 $\pm$ 1.00} &
\textbf{3.91 $\pm$ 2.58} &
\textbf{1.73 $\pm$ 1.04} &
\textbf{3.78 $\pm$ 2.72} &
\textbf{1.89 $\pm$ 1.11} &
\textbf{4.91 $\pm$ 2.66} &
\textbf{3.40 $\pm$ 1.06}  \\
				\bottomrule
			\end{tabular}}
		}
	\end{center}
	\label{table:general1}
\end{table*}
\modifylatest{\subsubsection{Generalization Analysis}
To validate the generalization of I2PNet, we conduct the generalization analysis on datasets collected on different experimental platforms, including nuScenes dataset~\cite{nuscenes2019} on a car-like platform and M2DGR dataset~\cite{yin2021m2dgr} on a ground robot platform. The model is trained on KITTI Odometry dataset~\cite{geiger2013vision} and tested on those datasets. We also conduct experiments on DeepI2P~\cite{li2021deepi2p} and CorrI2P~\cite{ren2022corri2p} and report their performances. The results are shown in Table~\ref{table:general1}, where "failed" indicates that there are no samples whose RREs are less than $10^\circ$ and RTEs are less than $5m$ as well.
Table~\ref{table:general1} shows that our proposed method achieves the best performance on both car-like platform and ground robot platform dataset. 
For the car-like platform generalization test on nuScenes dataset, although our method's accuracy slightly increases from an error of around $1^\circ$ and $0.5m$ to around $3^\circ$ and $ 2m $ in generalization tests, \emph{DeepI2P (3D)}~\cite{li2021deepi2p}, \emph{DeepI2P (2D)}~\cite{li2021deepi2p}, and CorrI2P~\cite{ren2022corri2p} all fail. \emph{Grid. Cls. + EPnP}~\cite{li2021deepi2p} exhibits some level of generalization, but the generalization results have significant errors, reaching around $5^\circ$ and $ 3m $. Therefore, among all the compared methods, our approach significantly outperforms existing methods in terms of the generalization performance on the car-like platform.
As for the ground robot platform test on M2DGR dataset, the majority of methods trained on car-like KITTI Odometry dataset fail to generalize. Only \emph{Grid. Cls. + EPnP}~\cite{li2021deepi2p} shows some level of generalization in outdoor, hall, and lift scenes. However, it fails in narrow indoor settings. In contrast, our method, trained on car-like autonomous driving platforms, demonstrates strong generalization across various scenes such as the outdoor, hall, lift, and room. The experiment results on the M2DGR dataset highlight that the 2D-3D localization method proposed in this paper surpasses previous research in generalization performance across different robot platforms.
}

\modify{\subsection{Small-Range Localization}
The methods adopting end-to-end 2D-2D registration architectures~\cite{cattaneo2019cmrnet,chang2021hypermap,chen2022i2d} utilize the LiDAR depth image as the network input. This limits the localization range of these methods. To compare I2PNet with these methods, the small-range localization task is conducted in this subsection.}

\subsubsection{Dataset and Data Pre-processing}\label{section: lo_data}
The experiments are conducted on the KITTI Odometry dataset~\cite{geiger2013vision}. As the previous methods~~\cite{cattaneo2019cmrnet,chang2021hypermap,chen2022i2d}, the 03, 05, 06, 07, 08, and 09 sequences of the dataset are selected as the training set, and the separate 00 sequence is selected as the test set. The global LiDAR map is built by the frame poses provided by a LiDAR-based SLAM system as CMRNet \cite{cattaneo2019cmrnet}. After gaining the global LiDAR point cloud map, we localize the robot with pose initialization $H_{init}$. In addition, we crop the local 3D LiDAR map around $H_{init}$ to limit the scale of the point cloud inputted into the network for efficient and precise feature extraction of the point cloud. Then, the local 3D LiDAR map is transformed by the pose initialization $H_{init}$ to the local map coordinate system. Thus, the localization task is to estimate the relative pose $\phi_e$ between the local map coordinate system and the robot coordinate system, in which $\phi_e$ is the residual pose between the pose initialization and the ground truth robot pose. We simulate $\phi_e$ by random transformation generation. Specifically, $\phi_e$ is a composition of the random rotation within $[-10^\circ,10^\circ]$ and random translation within $[-2m,2m]$ at each of the $x$, $y$, and $z$ axes, which is much smaller than the range in the large-range localization task. As for the image, the top 50 rows of each image are cropped. Then, we resize the cropped image into the size $384\times1280$ as the input image of the network. 

As for the input point cloud, we randomly sample $8192$ points from the local LiDAR map.
Notably, to fairly compare I2PNet with the 2D-2D registration architectures, the point cloud accumulation is performed to build the global LiDAR map. However, the spherical projection can only be performed on the point cloud obtained from a single scan of the rotating LiDAR. Thus, the spherical projection is not used in the experiment of this section. The vanilla neighborhood query and sampling of PointNet++ are adopted to extract the point features.  
Moreover, we set the initial point features $F^0$ as the feature matrix with the size of $8192\times 4$. In the feature matrix, each feature vector is the concatenation of a three-dimensional zero vector and the intensity.

\setlength{\tabcolsep}{3mm}
\begin{table*}[t]
\renewcommand{\arraystretch}{1}
		\caption{Small-Range Localization Error on KITTI Odometry Dataset. \modifynew{The Results in Parentheses Are Reproduced by Ourselves. The Reproduced Results of HyperMap Are Not Provided as HyperMap Does Not Release the Codes}}
	\begin{center}
		\resizebox{0.75\linewidth}{!}
		{
			\begin{tabular}{c||cc|cc}
				\toprule

				Method&
 Rot.~($^\circ$)~$\downarrow$&Transl.~(m)~$\downarrow$ & Median Rot.~($^\circ$)~$\downarrow$&Median Transl.~(m)~$\downarrow$ \\
    \hline
    \hline
    \noalign{\smallskip}

    CASELITZ~\cite{caselitz2016monocular} &1.65 $\pm$ 0.91&
 0.30 $\pm$ 0.11&---&---
    
   \\
    \hline
    \noalign{\smallskip}
   
				CMRNet~\cite{cattaneo2019cmrnet}  &
				\modifynew{--- (1.98 $\pm$ 1.30)} & \modifynew{--- (0.62 $\pm$ 0.43)} &1.39 \modifynew{(1.68)}&0.51 \modifynew{(0.51)}
    				\\CMRNet++~\cite{cattaneo2020cmrnet++}  
				&\modifynew{--- (1.88 $\pm$ 1.43)}  & \modifynew{--- (0.70 $\pm$ 0.48)}&1.46 \modifynew{(1.52)}&0.55 \modifynew{(0.58)}
    \\      HyperMap~\cite{chang2021hypermap}&---&---&1.42&0.48 
    				\\I2D-Loc~\cite{chen2022i2d} 
				&\modifynew{--- (1.07 $\pm$ 1.17)} &\modifynew{--- (0.34 $\pm$ 0.38)}&0.70 \modifynew{(0.77)}&0.18 \modifynew{(0.21)}

 				\\\midrule
    				Ours (I2PNet)
				&\textbf{0.74 $\pm$ 0.40} & \textbf{0.08 $\pm$ 0.06}&\textbf{0.67}&\textbf{0.07}
				\\
        		
				\bottomrule
			\end{tabular}
		}
	\end{center}
	\label{table:loresult}
\end{table*}

\begin{table*}[h]
\renewcommand{\arraystretch}{1.5}
\setlength{\tabcolsep}{3pt}
		\caption{\modifylatest{Generalization Experiments on KITTI Odometry, nuScenes, Argoverse, Waymo, and Lyft5 Datasets on Small-Range Localization Task. The Results of CMRNet, CMRNet++ and I2DLoc Are Reproduced by Ourselves for Extensive Generalization Testing}}

	\begin{center}
		\resizebox{1\linewidth}{!}
		{
\modifylatest{\begin{tabular}{c|c||cc|cc|cc|cc|cc}
\toprule 
\multirow{2}{*}{Training Dataset} &\multirow{2}{*}{Method} & \multicolumn{2}{c|}{KITTI} & \multicolumn{2}{c|}{nuScenes} & \multicolumn{2}{c|}{Argoverse} & \multicolumn{2}{c|}{Waymo} & \multicolumn{2}{c}{Lyft5}\\
& &Rot.~($^\circ$)~$\downarrow$ & Transl.~($m$)~$\downarrow$ 
&Rot.~($^\circ$)~$\downarrow$ & Transl.~($m$)~$\downarrow$
&Rot.~($^\circ$)~$\downarrow$ & Transl.~($m$)~$\downarrow$
&Rot.~($^\circ$)~$\downarrow$ & Transl.~($m$)~$\downarrow$
&Rot.~($^\circ$)~$\downarrow$ & Transl.~($m$)~$\downarrow$
\\
\hline
\hline
\noalign{\smallskip}
\multirow{4}{*}{KITTI}
&CMRNet~\cite{cattaneo2019cmrnet} & 1.98$\pm$1.30 & 0.62$\pm$0.43 & 7.12$\pm$3.20 & 1.67$\pm$0.64  & 9.49$\pm$3.09 & 1.65$\pm$0.77  & 11.24$\pm$3.30 & 2.01$\pm$0.85   & 7.97$\pm$2.96 & 1.83$\pm$0.61 
\\
&CMRNet++~\cite{cattaneo2020cmrnet++} & 1.88$\pm$1.43 & 0.70$\pm$0.48 & 7.70$\pm$4.40 & 1.61$\pm$0.78  & 3.92$\pm$2.44 & 1.33$\pm$0.69  & 8.84$\pm$3.91 & 1.88$\pm$0.83   & 7.96$\pm$3.95 & 1.70$\pm$0.77 
\\
& I2DLoc~\cite{chen2022i2d} & 1.07$\pm$1.17 & 0.34$\pm$0.38 & 6.60$\pm$5.33 & 1.46$\pm$0.81 & 3.09$\pm$2.45 & 1.12$\pm$0.70 & 7.17$\pm$5.35 & 1.87$\pm$0.82 & 7.95$\pm$4.99 & 1.52$\pm$0.73 \\
\cline{2-12}
& Ours (I2PNet) &\textbf{0.74$\pm$0.40} & \textbf{0.08$\pm$0.06}
&\textbf{2.88$\pm$0.93} & \textbf{1.25$\pm$0.53}
&\textbf{0.78$\pm$0.15} & \textbf{0.68$\pm$0.14}
&\textbf{0.90$\pm$0.55} & \textbf{0.62$\pm$0.28}
&\textbf{0.76$\pm$0.10} & \textbf{0.51$\pm$0.05}
\\
\midrule
\multirow{4}{*}{nuScenes}
& CMRNet~\cite{cattaneo2019cmrnet}  & 5.69$\pm$2.93 & 1.74$\pm$0.84  & 2.77$\pm$1.87 & 1.02$\pm$0.68 & 4.24$\pm$2.25 & 1.98$\pm$0.95  & 5.26$\pm$3.00 & 2.10$\pm$0.77   & 5.10$\pm$2.04 & 1.70$\pm$0.64 
\\
& CMRNet++~\cite{cattaneo2020cmrnet++} & 5.56$\pm$3.43 & 1.49$\pm$0.76 & 3.88$\pm$2.70 & 1.21$\pm$0.67 & 4.86$\pm$3.18 & 1.62$\pm$0.81  & 6.22$\pm$4.09 & 2.01$\pm$0.95   & 5.69$\pm$3.38 & 1.52$\pm$0.70
\\
& I2DLoc~\cite{chen2022i2d}  & 4.46$\pm$3.28 & 1.25$\pm$0.72 & 2.70$\pm$2.50 & 0.83$\pm$0.60 & 3.89$\pm$3.31 & 1.33$\pm$0.74 & 6.36$\pm$3.97 & 1.97$\pm$0.76 & 6.16$\pm$3.31 & 1.66$\pm$0.72 \\
 \cline{2-12}
& Ours (I2PNet) 
&\textbf{1.59$\pm$0.27} & \textbf{1.14$\pm$0.27}
&\textbf{0.57$\pm$0.75} & \textbf{0.58$\pm$0.42}
&\textbf{1.44$\pm$0.40} & \textbf{0.72$\pm$0.19}
&\textbf{1.25$\pm$0.67} & \textbf{0.72$\pm$0.31}
&\textbf{1.50$\pm$0.39} & \textbf{0.69$\pm$0.13}
\\
				\bottomrule
			\end{tabular}}
		}
	\end{center}
	\label{table:cmrnetgen}
\end{table*}

\subsubsection{Experiment Result and Visualization}\label{section:lo_result}
We utilize the Rotation angle (Rot.) and Translation length (Transl.) of the error between the final monocular camera pose estimation and the ground truth pose to evaluate the accuracy of the localization. Specifically, since the ground truth pose is $H_{gt}=\phi_{e}H_{init}$ and the final pose estimation is $H_{final}=\phi_{pred}H_{init}$, the error $H_e$ between $H_{gt}$ and $H_{final}$ is $H_{final}H_{gt}^{-1}=\phi_{pred}\phi_{e}^{-1}$. Thus, the formulas of the Rot. and Transl. are:
\modifynew{\begin{equation}
	Rot.=\arccos \frac{tr\left( R_{e} \right) -1}{2},
	Transl.=\lVert t_{e} \rVert _2,
\end{equation}}
in which $R_e$ and $t_e$ are the rotation matrix and translation vector of $H_e$ respectively. To validate the generalization of our model, we calculate the average median, mean, and standard deviation of the Rot. and Transl. results of ten experiments on the test set.

 In Table~\ref{table:loresult}, the performance of I2PNet is compared with the end-to-end 2D-2D registration-based methods~\cite{cattaneo2019cmrnet,chang2021hypermap,chen2022i2d}, using the above metrics. In addition, for complete comparison, the conventional method CASELITZ~\cite{caselitz2016monocular} and the RANSAC-based method CMRNet++~\cite{cattaneo2020cmrnet++}, are as well included in the comparison. 
The results in Table~\ref{table:loresult} indicate that our I2PNet has the smallest final pose estimation error. 
Compared to CASELITZ~\cite{caselitz2016monocular}, I2PNet only requires a single image rather than the local bundle adjustment reconstruction from image sequences since the 2D-3D cost volume module can directly match the pixels and points. 
Compared to CMRNet++~\cite{cattaneo2020cmrnet++}, I2PNet is an end-to-end architecture, while CMRNet++~\cite{cattaneo2020cmrnet++} adopts the separate RANSAC-based pose estimation module. As in the large-range localization task, I2PNet benefits from end-to-end optimization and thus has better localization accuracy.
\modifynew{Compared to the end-to-end 2D-2D registration-based methods~\cite{cattaneo2019cmrnet,chang2021hypermap,chen2022i2d}, I2PNet avoids projecting the 3D point cloud to a depth map and enables learning the correspondence between 3D structural features and 2D image features on a camera intrinsic-independent plane. The 2D depth map is an indirect representation of the point cloud and depends on the camera intrinsic, which would prevent direct extraction of 3D structural features from the raw point cloud.} 
Therefore, I2PNet using the 2D-3D registration performs better localization.

\modifylatest{
\subsubsection{Generalization Comparison}
The generalizability is also validated on the small-range localization. The models are not only trained and tested on KITTI Odometry~\cite{geiger2013vision} and nuScenes~\cite{nuscenes2019} dataset respectively, but also tested on three other vehicle datasets: Argoverse~\cite{chang2019argoverse}, Waymo~\cite{sun2020waymo}, and Lyft5~\cite{houston2021one} datasets. For training on the nuScenes dataset, We randomly selected 70 traversals from the trainval data for training and 13 traversals from the test data for testing. Then, we construct the global LiDAR map, crop the local LiDAR map, and generate the ground truth relative pose $\phi_{e}$ by random transformation generation on the nuScenes dataset as the KITTI Odometry dataset. For Argoverse dataset, the sequence train4 is used for generalization tests. For Waymo dataset, the sequence 00, 02, 03, 04, 05, and 07 in the validation part of the Perception Dataset is used as the test dataset. As for Lyft5 dataset, all the 10 urban scenes in the Perception Dataset is used for tests.
Tables~\ref{table:cmrnetgen} shows the experimental results of training on the KITTI Odometry and nuScenes datasets, and testing on all the 5 datasets. Since CMRNet~\cite{cattaneo2019cmrnet}, CMRNet++~\cite{cattaneo2020cmrnet++} and I2DLoc~\cite{chen2022i2d} do not provide generalization results under the same training and testing settings as ours, we only list the reproduced results.
The experimental results indicate that our proposed method achieves the best performance not only on the standard training and testing settings of the KITTI Odometry and nuScenes datasets individually but also in generalization tests on Argoverse, Waymo and Lyft5 datasets. Our method outperforms the compared methods, CMRNet~\cite{cattaneo2019cmrnet}, CMRNet++~\cite{cattaneo2020cmrnet++}, and I2DLoc~\cite{chen2022i2d}, by nearly 50\% on the performance both on training dataset's testing set and the datasets for generalization tests. This strongly demonstrates the effectiveness and generalization capability of our proposed method.}

\subsection{Ablation Studies}\label{section:ab}
In this subsection, we conduct ablation studies to discuss the effect of the input point number and the effectiveness of the proposed modules. Our ablation studies are all conducted on the large-range localization task with the KITTI Odometry dataset. 
\subsubsection{Effect of the Input Point Number}
In Section~\ref{section:fe}, we introduce the stride-based sampling and projection-aware grouping to replace FPS and neighborhood query among all the points in vanilla PointNet++. The replacement makes us able to process all the points in the raw point cloud efficiently. The network using all the points can fully utilize the information of the raw point cloud and thus has a better localization performance. To validate this, we compare the RRE/RTE localization error between the network using fixed-size points and the network using all the points. We also set different numbers of the input points as 8192, 16384, and 24576. 
The results in Table~\ref{table:points} show that the increment of input points number can effectively improve the localization performance. Moreover, the network using all the points has the best performance. 
In Section~\ref{section:ef}, we will show the network  using all the points is still efficient.

\setlength{\tabcolsep}{0.9mm}
\begin{table}[t]
		\caption{Ablation Study on the Number of Input Points}
	\begin{center}
		\resizebox{0.85\linewidth}{!}
		{
			\begin{tabular}{l||cc}
				\toprule 
				
				Method&
 RRE ($^\circ$)~$\downarrow$ & RTE (m)~$\downarrow$      \\
				\hline\hline
    \noalign{\smallskip}

				Ours (w/ $8192$ input points)    
				&2.16 $\pm$ 1.90 & 0.57 $\pm$ 0.37
				\\
				Ours (w/ $16384$ input points) 
				&2.05 $\pm$ 1.85 & 0.48 $\pm$ 0.36
				\\
				Ours (w/ $24576$ input points) & 
1.61 $\pm$ 1.42
 & 0.36  $\pm$ 0.29
    \\ Ours (w/ all input points)& \textbf{0.83 $\pm$ 1.04} &
\textbf{0.21 $\pm$ 0.29}\\
				\bottomrule
			\end{tabular}
		}
	\end{center}
	\label{table:points}
\end{table}
\setlength{\tabcolsep}{0.9mm}
\begin{table}[t]
	\centering

		\caption{Ablation Study on the Effectiveness of the Proposed Modules}
	
		\resizebox{1\linewidth}{!}
		{
			\begin{tabular}{l||cc}
				\toprule 
			
				Method&
 RRE ($^\circ$)~$\downarrow$ & RTE (m)~$\downarrow$      \\
			
				\hline\hline
    \noalign{\smallskip}
				Ours (w/o 2D-3D Cost Volume Module)   
				
&2.54 $\pm$ 1.94&3.29 $\pm$ 1.19
				\\
               \modifylatest{Ours (w/o Intrinsic-Independent Space)}   & \modifylatest{1.85 $\pm$ 1.58} &
\modifylatest{0.71 $\pm$ 0.56}
                \\
				Ours (w/o Outlier Mask)
&1.12 $\pm$ 1.28&
0.28 $\pm$ 0.27
			
				\\
				Ours (w/o Fine Registration) &
				
    1.59 $\pm$ 1.41&
0.60 $\pm$ 0.60
				\\
				Ours (w/o Pose Warping) &
         
            1.63 $\pm$ 1.41 &
0.53 $\pm$ 0.54
				\\
    				Ours (w/o \modify{PST Embedding}) 
				
    &1.01 $\pm$ 1.13 & 0.29 $\pm$ 0.33
				\\
			    Ours (I2PNet) &
\textbf{0.83 $\pm$ 1.04} &
\textbf{0.21 $\pm$ 0.29}
				\\
				\bottomrule
			\end{tabular}}
		
	\vspace{-7mm}
	\label{table:modules}
\end{table}
\setlength{\tabcolsep}{1mm}
\begin{table}[t]

		\caption{Efficiency Comparison}
  \vspace{-6mm}
	\begin{center}
		\resizebox{1.0\linewidth}{!}
		{
\begin{tabular}{l||cccc}
\toprule 
Method& Network size (MB) &  Inference (s)      \\
\hline\hline
\noalign{\smallskip}

Grid. Cls. + EPnP~\cite{li2021deepi2p} &100.75&0.051\\

DeepI2P (3D)~\cite{li2021deepi2p} &100.12&16.588\\ 
DeepI2P (2D)~\cite{li2021deepi2p} &100.12&9.388 \\ 
CorrI2P~\cite{ren2022corri2p} &141.07&2.984 \\
\midrule

Ours (I2PNet) (8192 w/o projection) & \textbf{3.38} &0.069\\
Ours (I2PNet) (16384 w/o projection) &\textbf{3.38} &0.143 \\
Ours (I2PNet) (24576 w/o projection) &\textbf{3.38} & 0.271\\
Ours (I2PNet) (all points w/ projection) &\textbf{3.38} & \textbf{0.048}\\
				\bottomrule
			\end{tabular}
		}
	\end{center}
	
	\label{table:efficiency}
\end{table}

\begin{table}[t]
    \centering
    \caption{\modifylatest{Large-Range Localization Error with Different Number of Iterations in the Fine Registration}}
    \setlength{\tabcolsep}{1mm}
    \resizebox{1\linewidth}{!}
	{
      \modifylatest{  \begin{tabular}{c|ccc|c}
\toprule

Iter Num & RRE ($^\circ$)~$\downarrow$ & RTE (m)~$\downarrow$ & Recall (\%)~$\uparrow$ & Inference Time (s)~$\downarrow$  \\

\hline
\noalign{\smallskip}
1 & 0.83 $\pm$ 1.04 & 0.21 $\pm$ 0.29 & 98.67  & 0.048 \\
2 & 0.79 $\pm$ 0.92 & 0.20 $\pm$ 0.25 & 99.14  & 0.056 \\
3 & 0.78 $\pm$ 0.86 & 0.20 $\pm$ 0.27 & 99.32  & 0.063 \\
4 & 0.77 $\pm$ 0.83 & 0.20 $\pm$ 0.27 & 99.28 & 0.069 \\
5 & 0.78 $\pm$ 0.83 & 0.20 $\pm$ 0.29 & 99.32  & 0.076 \\
6 & 0.81 $\pm$ 0.89 & 0.21 $\pm$ 0.31 & 99.43 & 0.083 \\
\bottomrule

\end{tabular}}
}
\label{tab:iter}
\end{table}

\subsubsection{Effectiveness of the Proposed Modules}
We also conduct ablation studies to show the effectiveness of the proposed modules. In the subsequent ablation studies, despite the mentioned differences, the modules and hyperparameters are the same as the proposed I2PNet.

 \textbf{Effectiveness of the proposed 2D-3D cost volume module.} In the ablation study, \emph{Ours (w/o 2D-3D Cost Volume Module)}, the image-to-point cloud attentive fusion module proposed by DeepI2P~\cite{li2021deepi2p} is used to replace the 2D-3D cost volume module. The attentive fusion module of DeepI2P performs the attentive aggregation of the image features for each point. Then, the aggregated point-wise image features are fused with the point features by concatenation.
	The quantitative results of \emph{Ours (w/o 2D-3D Cost Volume Module)} in Table \ref{table:modules} show that the localization performance decreases after replacing the 2D-3D cost volume module with a simple cross-modality fusion module. The decrement indicates that 2D-3D association is not a simple cross-modality feature fusion. Our 2D-3D cost volume module implicitly constructs the point-pixel correspondences and
	embeds essential position information for spatial transformation estimation. In contrast, the attentive fusion module just performs the cross-modality feature fusion. Thus,  the network using the simple attentive fusion module has a larger relative rotation error and can hardly learn to register the image and point cloud in the aspect of translation.

    \modifylatest{\textbf{Effectiveness of the intrinsic-independent space.} In the ablation study, \emph{Ours (w/o Intrinsic-Independent Space)}, the intrinsic-independent space is not adopted. Instead, the pixel plane is treated as the space for the 2D-3D cost volume estimation and pose regression. Since the intrinsic parameters varies in the training and testing sets of KITTI Odometry dataset, the localization performance decreases when using the pixel plane, as shown in Table \ref{table:modules}. This result indicates the intrinsic-independent space ensures the consistent projection coordination of the 3D points when using the data with different intrinsic parameters and thus enables I2PNet to more correctly learn the spatial transformation between the image and point cloud.
}
    
    \textbf{Effectiveness of the outlier mask prediction module.} In the ablation study, \emph{Ours (w/o Outlier Mask)}, 
	the global spatial transformation embedding feature is the simple average of the 2D-3D cost volumes rather than the weighted sum of the 2D-3D cost volumes with the weights of the outlier masks. The quantitative results of \emph{Ours (w/o Outlier Mask)} in Table \ref{table:modules} show that the learned outlier masks can effectively filter the outliers and thus result in a smaller localization error.

    \textbf{Effectiveness of the fine registration.} In this ablation study, \emph{Ours (w/o Fine Registration)}, 
	only the coarse registration is performed.
	The quantitative results of \emph{Ours (w/o Fine Registration)} in Table \ref{table:modules} show that the coarse-to-fine registration architecture results in better registration performance. More correct correspondences can be found in the refined image-point cloud pair and thus make the predicted relative pose more accurate.

    \textbf{Effectiveness of the pose warping.} In the ablation study, \emph{Ours (w/o Pose Wraping)}, we do not warp the point cloud before estimating the 2D-3D cost volumes in the fine registration. 
	The quantitative results of \emph{Ours (w/o Pose Warping)} in Table \ref{table:modules} show that pose warping improves the registration performance since it generates an image-point cloud pair with smaller misalignment.

    \textbf{Effectiveness of the \modify{PST} Embedding.} In the ablation study, \modify{\emph{Ours (w/o PST Embedding)}}, we directly output the implicit correspondence features as the 2D-3D cost volumes. The quantitative results of \modify{\emph{Ours (w/o PST Embedding)}} in Table \ref{table:modules} show that 
	the embedded spatial transformation improves the outlier mask prediction. 
Therefore, the localization is more accurate.

\setlength{\tabcolsep}{2.5mm}
\begin{table*}[t]
    \centering
    \caption{Online Calibration Results on the KITTI Raw Dataset}
    \resizebox{0.85\textwidth}{!}
	{
    \begin{tabular}{c||cc|cc|cc|cc}
\toprule
\multirow{2}{*}{\begin{tabular}{c}Method \end{tabular}} & \multicolumn{2}{c|}{T1} & \multicolumn{2}{c|}{T2a}&\multicolumn{2}{c|}{T2b}&\multicolumn{2}{c}{T3} \\
&MSEE~$\downarrow$ & MRR~$\uparrow$ & MSEE~$\downarrow$ & MRR~$\uparrow$ &MSEE~$\downarrow$ & MRR~$\uparrow$&MSEE~$\downarrow$ & MRR~$\uparrow$ \\
\hline \hline
\noalign{\smallskip}
$\beta$-RegNet~\cite{schneider2017regnet}&0.0480&53.23\%&0.0440&37.08\%&0.0460&34.14\%&0.0920&-1.89\%\\
TAYLOR~\cite{taylor2016motion}&-&-&-&-&-&-&0.0100&-\\
CalibNet~\cite{iyer2018calibnet}  &-&-&0.0220&-&0.0220&-&-&- \\
RGGNet~\cite{yuan2020rggnet}& 0.0210&78.40\%&0.0140&75.61\%&0.0170&72.64\%&0.0100&83.22\%\\
\midrule
Ours (I2PNet) &\textbf{0.00096}&\textbf{99.04\%} &\textbf{0.00084}&\textbf{98.61\%}&\textbf{0.00115}&\textbf{97.81\%}&\textbf{0.00154}&\textbf{97.21\%}     \\              
\bottomrule
\end{tabular}
}
    \label{tab:oc_result}
    \vspace{-3mm}
\end{table*}
\begin{figure*}[t]
	\centering
	\vspace{0mm}
	\resizebox{1.0\textwidth}{!}
	{
		\includegraphics[scale=1.0]{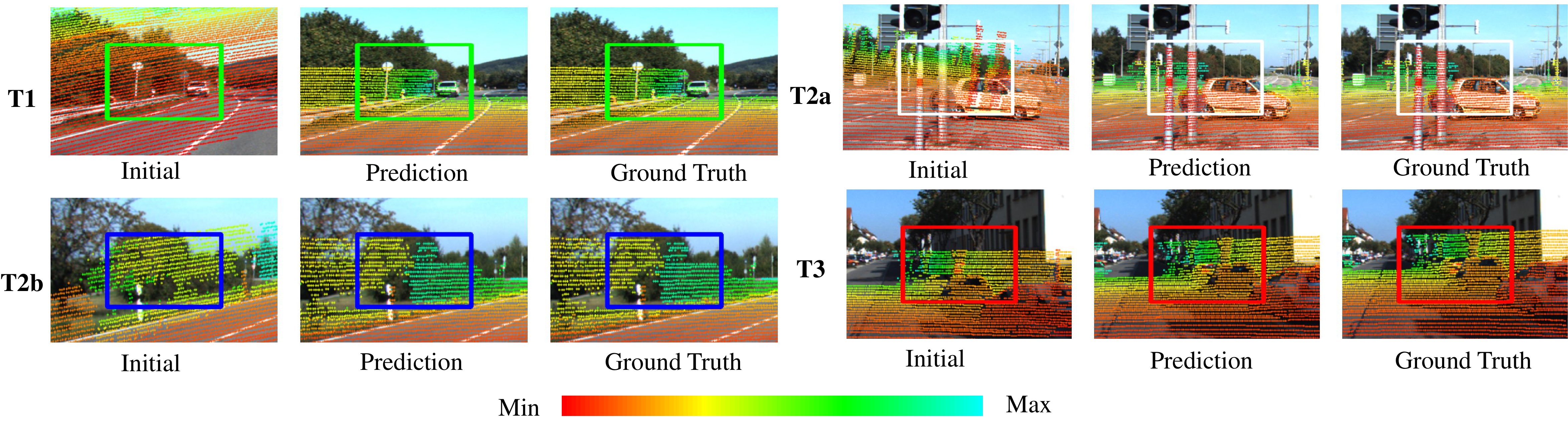}}
	\vspace{-3mm}
	\caption{Visualization of online calibration results. This figure presents the online calibration results on the four test sets. The pictures are cropped for better presentation and the areas that can reflect the calibration quality are marked by the bounding boxes. The color bar has the same meaning as Fig. \ref{fig:kd_vis}.
	}
	\vspace{-5mm}
	\label{fig:oc_vis}
	\end{figure*}

\subsection{Efficiency Evaluation}\label{section:ef}
In this subsection, we conduct the efficiency evaluation to validate the efficiency improvement of I2PNet to the previous works by the end-to-end 2D-3D registration architecture for the large-range robot localization. We present the evaluation results in Table~\ref{table:efficiency}. The results of \emph{Grid. Cls. + EPnP}, \emph{DeepI2P (3D)}, \emph{DeepI2P (2D)}, CorrI2P, and the I2PNets with different numbers of input points as the setting in Section~\ref{section:ab} are presented. 
For the former four methods, the inference time includes the pose estimation post-processing time evaluated on Intel(R) Xeon(R) Gold 6346 CPU and network inference time evaluated on an NVIDIA GeForce
RTX 3090. I2PNet does not need the pose estimation post-processing. Thus, the inference time only includes the network inference time.
From the efficiency comparison between the I2PNets with different numbers of input points in Table~\ref{table:efficiency}, we can conclude that the efficiency of the I2PNets with the vanilla sampling and neighborhood query methods decreases when the number of points increases. In contrast, by using efficient stride-based sampling and projection-aware grouping to replace the vanilla sampling and neighborhood query methods, the final proposed I2PNet using all input points has better efficiency than the I2PNets using fixed-size input points. Moreover, the final proposed I2PNet also has better efficiency than all the previous works. This indicates that our proposed end-to-end structure performs more accurate and efficient robot localization based on the camera image in the LiDAR point cloud map compared to the previous methods with the separate pose estimation module.

\subsection{\modifylatest{Iterations of the Fine Registration}}\label{sec:itera_fine}
\modifylatest{
To show the effect of the different iteration numbers in the fine registration, the experiments of different iteration numbers are conducted.  
 As shown in Table~\ref{tab:iter}, increasing the iteration count from 2 to 4 improves accuracy; beyond 4 iterations, the best performance is reached, and further iterations do not yield additional improvements, although they slightly increase recall. However, increasing number of iterations also reduces computational efficiency. As one iteration already achieves state-of-the-art performance with the highest efficiency compared to previous methods, we choose one iteration in the fine registration in our comparison experiments with other methods. In practice, the number of iterations can be adjusted based on the requirements of the localization accuracy and efficiency.}

\section{Extension to Camera-LiDAR Online Calibration}
In this section, we demonstrate the effectiveness of I2PNet when extended to the camera-LiDAR online calibration task.
\subsection{Dataset and Data Pre-processing}\label{section: oc_data}
Like the recent camera-LiDAR online calibration methods~\cite{taylor2016motion,schneider2017regnet,iyer2018calibnet,yuan2020rggnet}, the KITTI raw dataset~\cite{geiger2013vision} is adopted to evaluate our network. The same training set and test set are chosen as the recent methods. The ground truth extrinsic calibration matrix between each camera-LiDAR pair $H_{gt}$ is first calculated with the calibration data provided by the KITTI raw dataset. We adopt the method as RGGNet~\cite{yuan2020rggnet} to simulate the calibration error during the running of the robot. Specifically, $H_{gt}$ is randomly varied to obtain the initial calibration matrix $H_{initial}$ with calibration error, as follows:
\begin{equation}
   H_{initial} = \phi H_{gt},
\end{equation}
where $\phi$ is a random transformation matrix with the uniformly sampled translation error within the range of $\pm\gamma m$ in each translation axis and uniformly sampled rotation error within the range of $\pm \beta ^\circ$ in each rotation axis. Then, we use the initial calibration matrix $H_{initial}$ to multiply the original coordinates of the LiDAR points to obtain the LiDAR point cloud with the calibration error. Because $H_{gt}$ is the ground truth calibration matrix, $\phi$ is exactly the calibration error between the new point cloud and the image. Therefore, our network is to estimate the ground truth decalibration matrix $\phi_{gt}$, which is exactly the relative pose between the image and the new point cloud. $\phi_{gt}$ is calculated as follows:
\begin{equation}
   \phi_{gt}=\phi^{-1}.
\end{equation}
 $\phi_{gt}$ is transformed to the equivalent quaternion $q_{gt}$ and translation vector $t_{gt}$. 
By the dataset generation method, we generate the training set from all the drives except the 0005 and 0070 drives in the 09/26/2011 sequence of KITTI raw dataset. For a fair comparison with the recent methods, the training set consists of the following three subsets randomly sampled from the whole data: (1) 24000 samples with the calibration error range $(\pm 0.2m, \pm 15^\circ)$ for the training on both the rotation and the translation errors; (2) 3000 samples with the calibration error range $(\pm 0.3m, \pm 0^\circ)$ for the training on the pure translation error; (3) 3000 samples with error $(\pm 0m, \pm 20^\circ)$ for the training on the pure rotation error. For the test set,  the four different test sets are generated as well for a fair comparison: (1) \textbf{T1} consists of 2000 samples randomly sampled from the 0005 and 0070 drives in 09/26/2011 sequence with the calibration error range $(\pm 0.2m, \pm 15^\circ)$; (2) \textbf{T2a} consists of 2000 samples randomly sampled from all the drives except the 0005 and 0070 drives in 09/26/2011 sequence with the calibration error range $(\pm 0.2m, \pm 10^\circ)$; (3) \textbf{T2b} consists of 2000 samples randomly sampled from the 0005 and 0070 drives in 09/26/2011 sequence with the calibration error range $(\pm 0.2m, \pm 10^\circ)$; (4) \textbf{T3} consists of 2000 samples randomly sampled from the 0027 drive in 10/03/2011 sequence with the calibration error range $(\pm 0.3m, \pm 2^\circ)$ for the comparison with the conventional methods~\cite{taylor2016motion}. It is noticed that RGGNet~\cite{yuan2020rggnet} utilizes the independently sampled 2000 samples with the calibration error range $(\pm 0.3m, \pm 2^\circ)$ on the 0027 drive in 10/03/2011 sequence to finetune the model.
In contrast, we do not finetune our model on the extra training data.
In addition, we resize the input image to $352\times1216$ and input all the points in the raw point cloud. For the calculation of the 2D spherical coordinates, the initial upper bounds $(H,W)$ are $(64,1800)$. Up and down vertical field-of-views are $f_{up}=2.0,f_{down}=24.8$. Moreover, each feature vector of the initial point features $F^0$ is the concatenation of a three-dimensional zero vector and the intensity.

\subsection{Experiment Result and Visualization}
Table \ref{tab:oc_result} presents the quantitative experiment results between the I2PNet and the other works on the four test sets. 
$\beta$-RegNet is the re-implemented RegNet~\cite{schneider2017regnet} by RGGNet~\cite{yuan2020rggnet} which uses the network architecture of RGGNet~\cite{yuan2020rggnet} and the loss in RegNet~\cite{schneider2017regnet}.
In the experiment results, we utilize the same evaluation metrics as RGGNet~\cite{yuan2020rggnet}: mean $se3$ error (MSEE) and mean re-calibration rate (MRR). MSEE and MRR are based on the $se3$ distance between the predicted decalibration matrix $\phi_{pred}$ (which is calculated using the predicted $q_3$ and $t_3$) and ground truth decalibration matrix $\phi_{gt}$. The formulas of the metrics are:
\modifynew{\begin{equation}
MSEE = \frac{1}{n}\sum_{i=1}^{n} E_i,
MRR =  \frac{1}{n}\sum_{i=1}^{n} \frac{\eta_i-E_i}{\eta_i},
\vspace{-2mm}
\end{equation}}
where $E_i$ is the $se3$ distance between $\phi_{pred}$ and $\phi_{gt}$ of the $i$-th sample, $n$ is the number of samples, and $\eta_i$ is the miscalibration noise of the $i$-th sample.

As shown in Table~\ref{tab:oc_result}, our I2PNet has better MSEEs and MRRs than all the state-of-the-art works in all the test sets by a large margin. The results show that I2PNet is effectively extended to the online calibration task with high calibration accuracy. This indicates that the end-to-end 2D-3D registration architecture enables the wider application of I2PNet in various robot tasks and results in better registration.
In addition, it is noticed that on the T3 test set, I2PNet has better performance than the finetuned RGGNet and the traditional method TAYLOR without finetuning. The performance on the T3 test set shows that I2PNet is robust to various initial noise ranges. 

Fig.~\ref{fig:oc_vis} qualitatively shows the online calibration performance of I2PNet. The visualization shows that the initial misalignments on T1, T2a, and T2b test sets are large, while the initial misalignment on the T3 test set is small. Despite different initial misalignments, the predictions made by I2PNet have few differences from the ground truths on all four test sets. The results further demonstrate the effectiveness and generality of I2PNet on the camera-LiDAR online calibration task.

\section{Conclusion} \label{section:con}
In this paper, we introduced a novel image-to-point cloud registration architecture, I2PNet, for vehicle localization. 
I2PNet performs both high accuracy and efficient large-range image-based robot localization in the LiDAR point cloud map based on the end-to-end 2D-3D registration.  
We realized the end-to-end 2D-3D registration by the novel 2D-3D cost volume module and outlier mask prediction module. 
The end-to-end 2D-3D registration enables each module to be optimized by the united target. In addition, the complete 3D point cloud structure is preserved for the image-to-point cloud association in the architecture. Therefore, better registration accuracy is realized.

We conducted extensive experiments on multiple datasets and tasks to demonstrate the state-of-the-art camera localization ability of I2PNet in the LiDAR point cloud map. 
I2PNet can reach $0.83^\circ$ average RRE and $0.21m$ average RTE 
within a large localization range of $360^\circ$ and $10m$ on the KITTI Odometry dataset, improving $60.0\%$ average RRE and $71.6\%$ average RTE than the previous state-of-the-art methods. In addition, the performance of the same task on nuScenes dataset as well exceeds previous methods. 
\modifylatest{Moreover, based on the end-to-end architecture, the efficiency of I2PNet reaches 20Hz, outperforming the previous methods.}
I2PNet also outperforms the previous end-to-end 2D-2D registration-based methods in the small-range localization task. The median rotation error and median translation error of I2PNet are $0.67^\circ$ and $0.07m$, improving the best $0.70^\circ$ median rotation error and $0.18m$ median translation error of the previous methods by $4.3\%$ and $61.1\%$ respectively. We also performed generalization test on various datasets and demonstrated that I2PNet has better generalization ability than all previous methods on both two tasks.
As for the extension of I2PNet to camera-LiDAR online calibration, I2PNet reaches $98.17\%$ average re-calibration rate, exceeding the best $77.47\%$ average re-calibration rate of previous online calibration methods by $26.7\%$.

Finally, for our limitation and future work, our I2PNet can not directly handle the point cloud maps of sizes up to a few kilometers for global localization, since our approach performs localization through image-to-point cloud registration, which depends on the fine-grained feature extraction of the point cloud. However, it is feasible to integrate an image-to-point cloud place recognition module~\cite{cattaneo2020global, zheng2023i2p, shubodh2024lip} to obtain the coarse location of the robot within the global map, thereby constraining the size of the local point cloud map registered with the images. 
We put the integration with the image-to-point cloud place recognition module as future work.

\appendix  


\subsection{\modifylatest{Network Hyperparameters}}
\modifylatest{
The necessary hyperparameters of the modules in I2PNet are listed in Table~\ref{table:parameters}. 
$(S_h,S_w)$ are the strides of the stride-based sampling in the point cloud feature extraction. In the image feature extraction, $(S_h,S_w)$ are the strides of convolutional layers.
Notably, since the initial verticle upper bounds $H$ are set as different values for different LiDARs, the stride $S_h$ of the first point cloud feature extraction layer is set as 2 or 4 when $H$ is set as 32 or 64. $K$ is the number of nearest neighbors. Notably, the first $K$ of 2D-3D cost volumes in the coarse registration is set as the number of pixels in the third layer $M^3$ or 32, which represents the all-to-all point-pixel mixture or KNN-based point-pixel mixture respectively. Kernel size is the size of the 2D fixed-size kernels in the projection-aware grouping, while distance is the distance threshold in the projection-aware grouping. In image feature extraction, channel dimensions are the output channel dimensions of the convolutional layers. In the other modules, channel dimensions are the output channel dimensions of the shared MLP blocks or FC layers. 
}
 \begin{table*}[t]

	\caption{Network Hyperparameters of I2PNet}
	\begin{center}
		\centering
		\footnotesize
		\setlength{\tabcolsep}{0.5mm}
			\begin{center}
				\resizebox{0.9\textwidth}{!}
				{
					\begin{tabular}{@{}cccccccccc@{}}
						\toprule
						Module & \multicolumn{2}{c}{Layer Type} &$K$ &$(S_h,S_w)$ & Kernel Size& Distance&Channel Dimensions \\ 
						\midrule
      
						&\multicolumn{2}{c}{Layer 1} & --- &(4,4)&---&---& [16,16,16,16,32] \\
						&\multicolumn{2}{c}{Layer 2} & --- &(4,4)&---&---& [32,32,32,32,64] \\
						\multirow{-3}{*}{\begin{tabular}[c]{@{}c@{}}Image Feature Extraction\end{tabular}}&\multicolumn{2}{c}{Layer 3} & --- &(2,2)&---&---& [64,64,64,64,128] \\
						\cline{1-8}\noalign{\smallskip}

						&\multicolumn{2}{c}{Layer 1} & 32 &(2 or 4,8)&(9,15)&0.75 & [16,16,32]  \\
						&\multicolumn{2}{c}{Layer 2} & 16 &(2,2)&(9,15)&3.00  & [32,32,64]\\
						&\multicolumn{2}{c}{Layer 3} & 16 &(2,2)&(5,9)&6.00 & [64,64,128] \\
						&\multicolumn{2}{c}{Layer 4}& 16  &(1,2)&(5,9)&12.0 & [128,128,256]  \\
						\multirow{-5}{*}{\begin{tabular}[c]{@{}c@{}}Point Cloud \\Feature Extraction \end{tabular}}&\multicolumn{2}{c}{Context Gathering for $E^c$} & 16 &(1,2)&(5,9)&12.0& [128,64,64] \\
						
						\cline{1-8}\noalign{\smallskip}
						&\multicolumn{2}{c}{\multirow{1}{*}{Layer 4 for $E^c$}}  &  \multirow{1}{*}{$M^3$ or 32, 4}  & \multirow{1}{*}---& \multirow{1}{*}(3,5) & \multirow{1}{*}4.50&\multirow{1}{*}{[128,64,64], [128,64], [64]}  \\
						\multirow{-2}{*}{\begin{tabular}[c]{@{}c@{}}2D-3D Cost Volume \end{tabular}}
						&\multicolumn{2}{c}{\multirow{1}{*}{Layer 3 for $E^f$}}  &  \multirow{1}{*}{32, 4}  & \multirow{1}{*}---& \multirow{1}{*}(3,5)& \multirow{1}{*}4.50 & \multirow{1}{*}{[128,64,64], [128,64], [64]} \\
						
						\cline{1-8}\noalign{\smallskip}
						
						&\multicolumn{2}{c}{Upsampling Layer for $UE$}    & 8   & (1,1/2)&(5,9)&9.00&[128,64],[64] \\
						
						&\multicolumn{2}{c}{Upsampling Layer for $UM$}    & 8 & (1,1/2)&(5,9)&9.00&[128,64],[64] \\
						\cline{1-8}\noalign{\smallskip}						\multirow{-4.8}{*}{\begin{tabular}[c]{@{}c@{}}Upsampling Layer \end{tabular}}

						&\multicolumn{2}{c}{Cost Volume Optimization for $OE$}    & ---   & --- &---&---& [128,64] \\
						&\multicolumn{2}{c}{Outlier Mask Prediction for $M^c$}    & --- &---&---&---& [128,64] \\
						&\multicolumn{2}{c}{Outlier Mask Prediction for $M^f$}    & ---  &---&---&---& [128,64]  \\
						
						\cline{1-8}\noalign{\smallskip}
						\multirow{-5.5}{*}{\begin{tabular}[c]{@{}c@{}}Cost Volume Optimization \\ and Outlier Mask Prediction
						\end{tabular}}
						
						
						&\multicolumn{2}{c}{FC for Middle Feature}  & --- & --- &---&---& [256]\\
						&\multicolumn{2}{c}{FC for $q_c$, FC for $t_c$}  & --- & --- &---&---&  [4], [3]\\
						&\multicolumn{2}{c}{FC for $q_f^{i+1}$, FC for $t_f^{i+1}$}  & --- & --- &---&---&  [4], [3]  \\
						\bottomrule
						\multicolumn{1}{c}{\multirow{-5.5}{*}{\begin{tabular}[c]{@{}c@{}}Pose Regression \end{tabular}}}
				\end{tabular}}
			\end{center}
		\label{table:parameters}
		\vspace{-5mm}
	\end{center}
	\end{table*}

\ifCLASSOPTIONcaptionsoff
  \newpage
\fi

\bibliographystyle{IEEEtran}  
\bibliography{IEEEabrv, references} 
\begin{IEEEbiography}[{\includegraphics[width=1in,height=1.25in,clip,keepaspectratio]{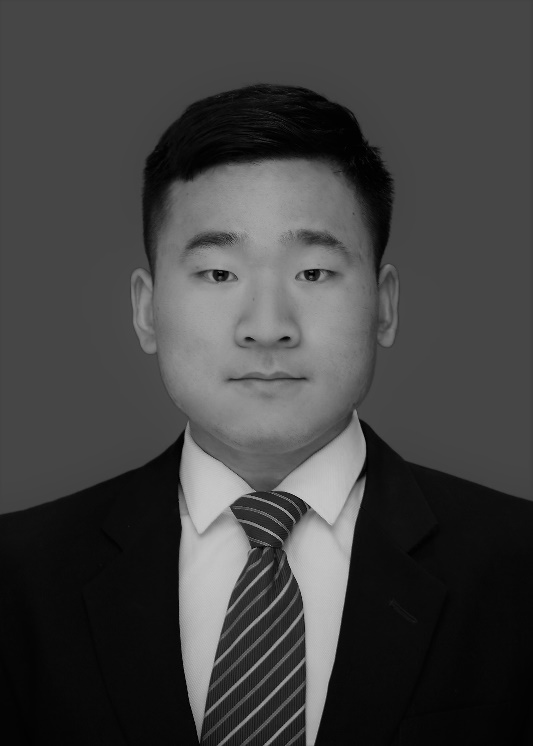}}]{Guangming Wang} received the B.S. degree from Department of Automation from Central South University, Changsha, China, in 2018, and Ph.D. degree in Control Science and Engineering from Shanghai Jiao Tong University, Shanghai, China. He visited Department of Computer Science of ETH Zurich from 2022 to 2023. He is currently a Research Associate with the Department of Engineering, University of Cambridge, UK. His current research interests include robot perception, localization, and mapping.
\end{IEEEbiography}
\begin{IEEEbiography}[{\includegraphics[width=0.9in,height=1.25in,clip,keepaspectratio]{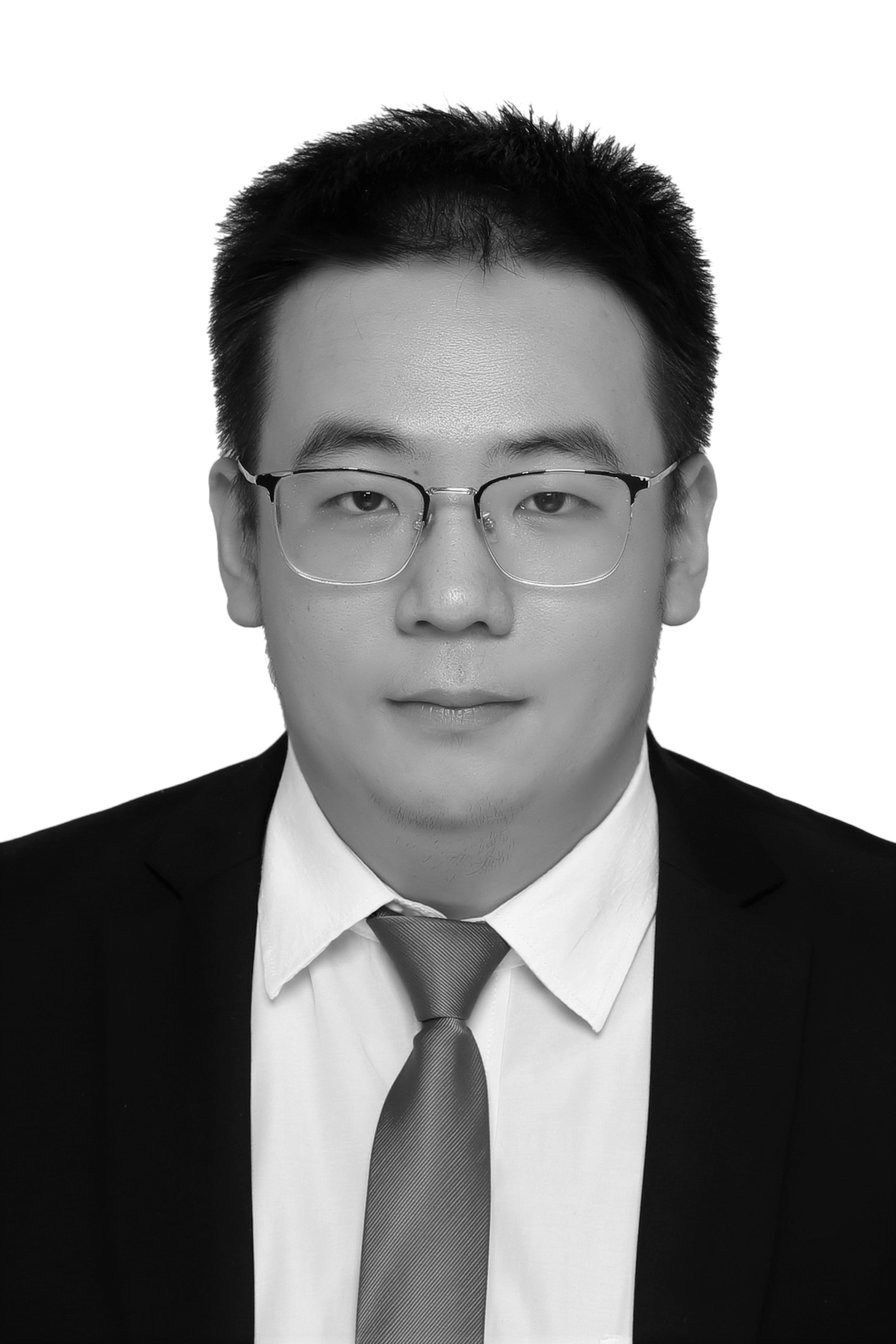}}]{Yu Zheng}
received the B.Eng degree in Department of Artificial Intelligence, Shanghai Jiao Tong University, China, in 2023. He is currently pursuing the Ph.D. degree in Control Science and Engineering with Shanghai Jiao Tong University, China. His current research interests include SLAM and computer vision.
\end{IEEEbiography}
\begin{IEEEbiography}[{\includegraphics[width=1in,height=1in,clip,keepaspectratio]{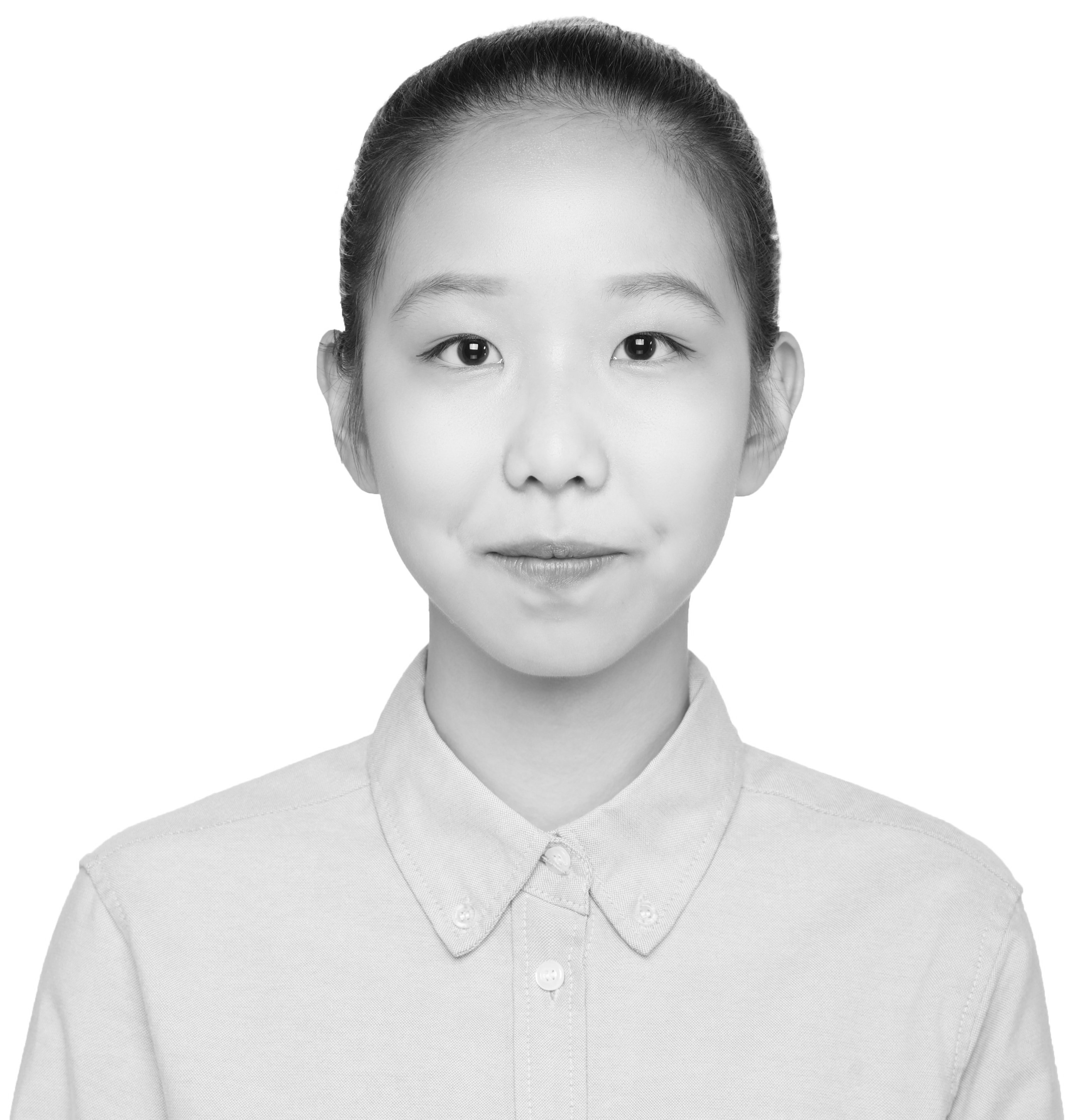}}]{Yuxuan Wu} received the B.Eng degree in School of Astronautics, Beihang University. He is currently pursuing the Ph.D. degree in Control Science and Engineering with Shanghai Jiao Tong University, China. Her latest research interests include SLAM and computer vision.
\end{IEEEbiography}
\begin{IEEEbiography}[{\includegraphics[width=0.9in,height=1.25in,clip,keepaspectratio]{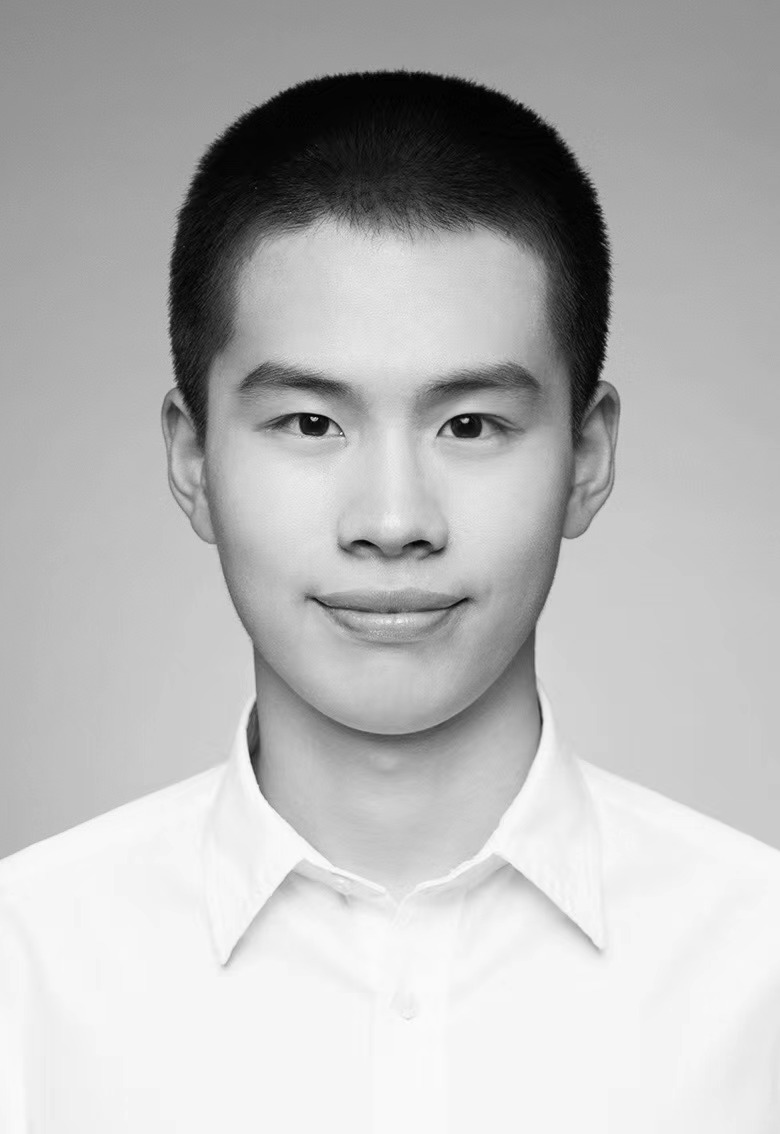}}]{Yanfeng Guo}
received his B.Eng degree at Department of Automation from Shanghai Jiao Tong University, Shanghai, China, in 2022. He is currently pursuing the M.S degree in Electrical and Computer Engineering at University of California, Los Angeles, the United States. His current research interests include computer vision and autonomous driving system.
\end{IEEEbiography}
\begin{IEEEbiography}[{\includegraphics[width=1in,height=1.25in,clip,keepaspectratio]{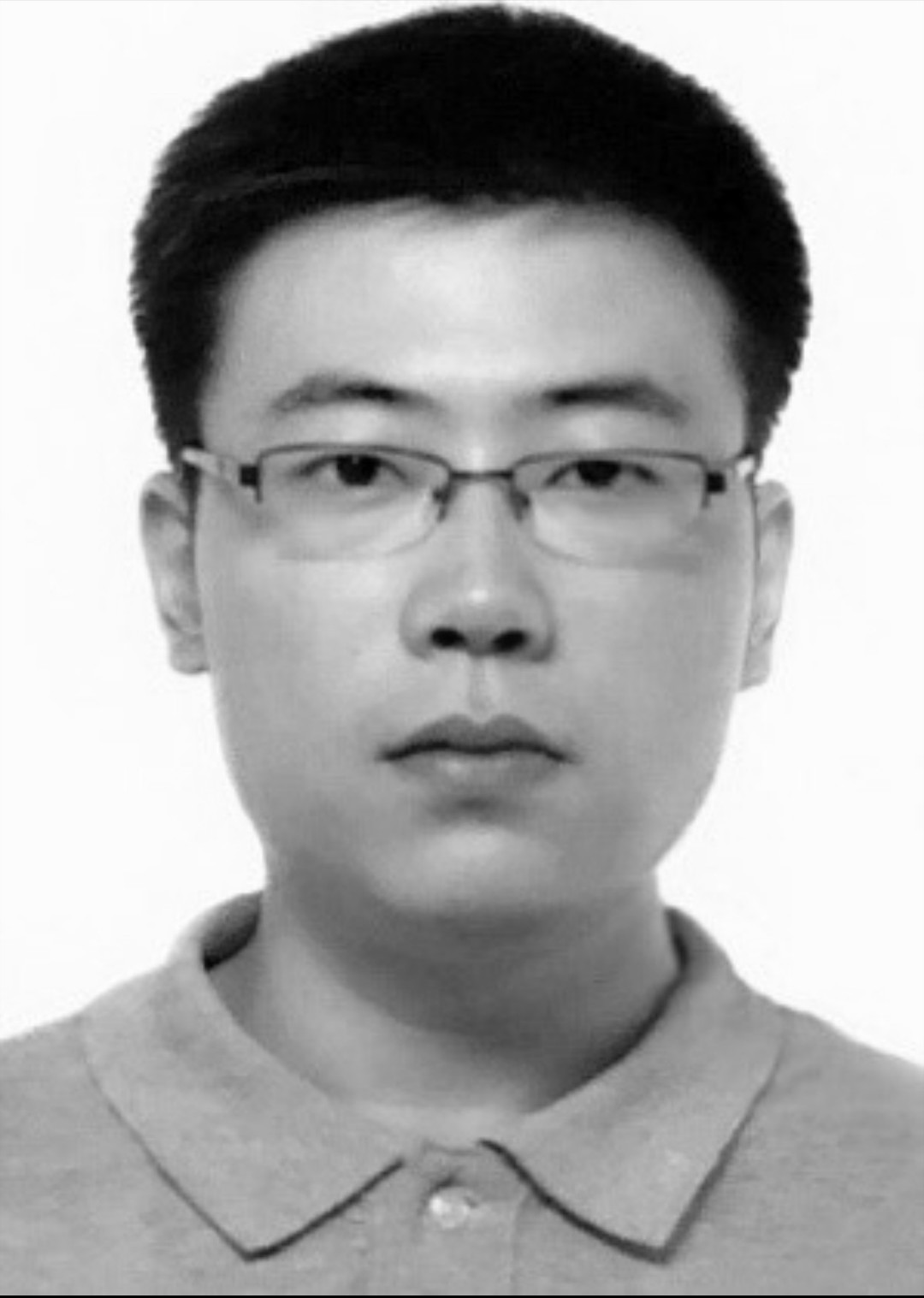}}]{Zhe Liu} received his B.S. degree in Automation from Tianjin University, Tianjin, China, in 2010, and Ph.D. degree in Control Technology and Control Engineering from Shanghai Jiao Tong University, Shanghai, China, in 2016. From 2017 to 2020, he was a Post-Doctoral Fellow with the Department of Mechanical and Automation Engineering, The Chinese University of Hong Kong, Hong Kong. From 2020 to 2022, he was a Research Associate of Department of Computer Science and Technology, University of Cambridge, United Kingdom. He is currently a tenure-track associate professor of Institute of Artificial Intelligence, Shanghai Jiao Tong University. His research interests include scheduling and optimization of large-scale robotic systems, resilient navigation of industrial self-driving systems, localization of self-driving systems, coordination of multi-robot formations, and robot control.
\end{IEEEbiography}
 \begin{IEEEbiography}[{\includegraphics[width=1.0in,height=1.25in,clip]{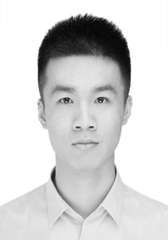}}]{Yixiang Zhu}  
received his B.Eng degree at Department of Electrical Engineering from Shanghai Jiao Tong University, Shanghai, China, in 2023. He is currently pursuing the M.S degree in Computer Control and Automation at Nanyang Technological University, Singapore. His current research interests include SLAM and computer vision.
\end{IEEEbiography}
\begin{IEEEbiography}[{\includegraphics[width=1in,height=1.25in,clip,keepaspectratio]{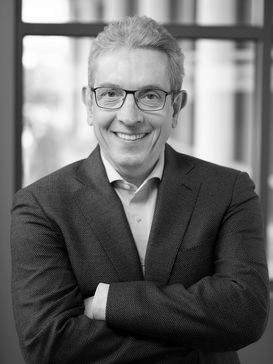}}]{Wolfram Burgard} 
is Vice President for Automated
Driving Technology at the Toyota Research Institute
in Los Altos, USA. He is on leave from a Professorship for Computer Science at the University of
Freiburg, Germany where he heads the Laboratory
for Autonomous Intelligent Systems. He received his
Ph.D. degree in computer science from the University
of Bonn in 1991. His areas of interest lie in robotics
and artificial intelligence. In the past, Wolfram
Burgard and his group developed several innovative
probabilistic techniques for robot navigation and
control. They cover different aspects including localization, mapping, path
planning, and exploration. For his work, Wolfram Burgard received several
best paper awards from outstanding national and international conferences.
In 2009, Wolfram Burgard received the Gottfried Wilhelm Leibniz Prize, the
most prestigious German research award. In 2010 he received the Advanced
Grant of the European Research Council. Wolfram Burgard is the spokesperson
of the Cluster of Excellence BrainLinks-BrainTools, President of the IEEE
Robotics and Automation Society, and fellow of the AAAI, EurAi and IEEE.
\end{IEEEbiography}
\begin{IEEEbiography}[{\includegraphics[width=1in,height=1.25in,clip,keepaspectratio]{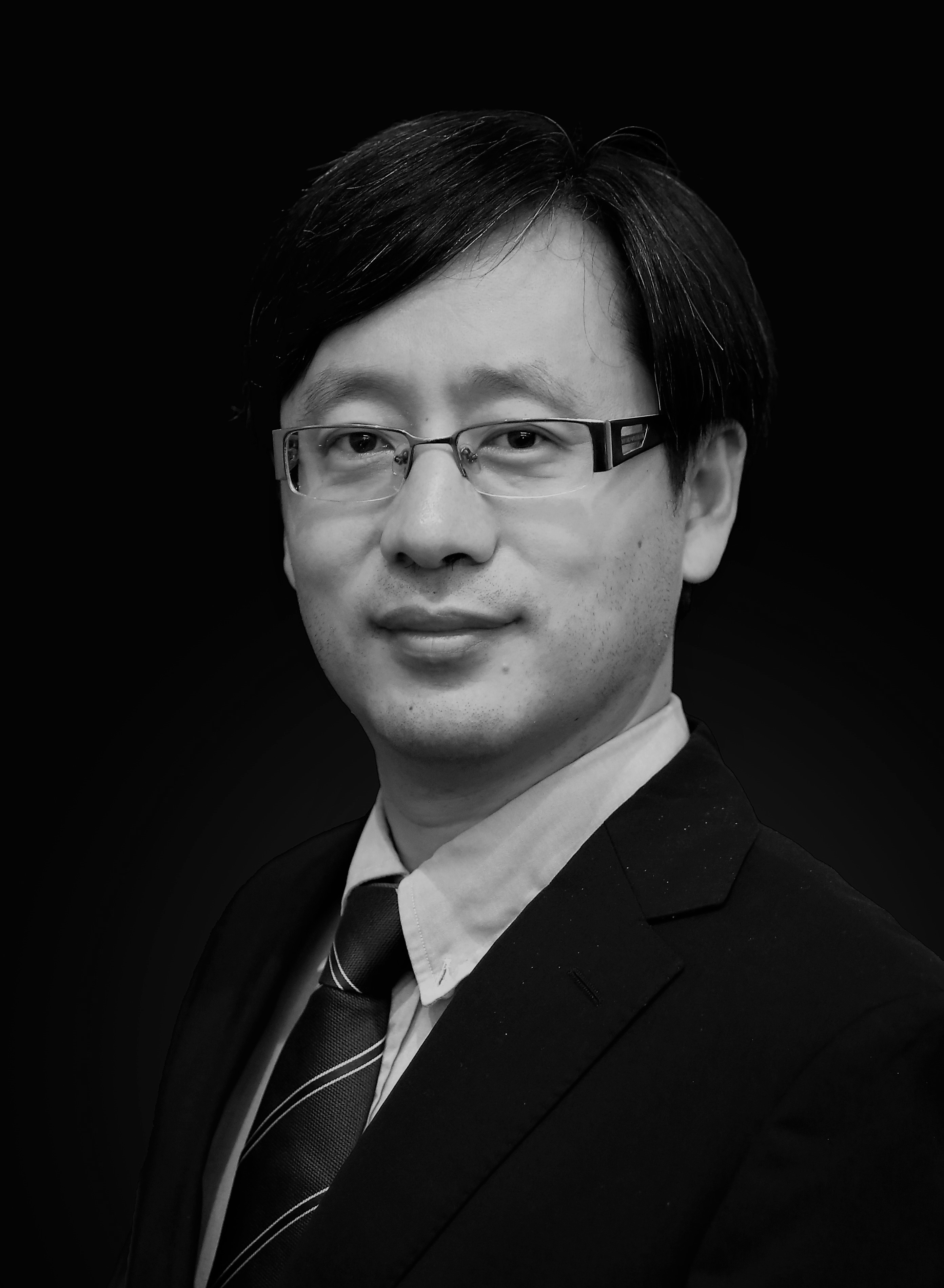}}]{Hesheng Wang}
  received the B.Eng. degree in electrical engineering from the Harbin Institute of Technology, Harbin, China, in 2002, and the M.Phil. and Ph.D. degrees in automation and computer-aided engineering from The Chinese University of Hong Kong, Hong Kong, in 2004 and 2007, respectively. He is currently a Professor with the Department of Automation, Shanghai Jiao Tong University, Shanghai, China. His current research interests include visual servoing, service robot, computer vision, and autonomous driving. 
Dr. Wang is an Associate Editor of IEEE Transactions on Automation Science and Engineering, IEEE Robotics and Automation Letters, Assembly Automation and the International Journal of Humanoid Robotics, a Technical Editor of the IEEE/ASME Transactions on Mechatronics, an Editor of Conference Editorial Board (CEB) of IEEE Robotics and Automation Society. He served as an Associate Editor of the IEEE Transactions on Robotics from 2015 to 2019. He was the General Chair of IEEE ROBIO 2022 and IEEE RCAR 2016, and the Program Chair of the IEEE ROBIO 2014 and IEEE/ASME AIM 2019. He will be the General Chair of IEEE/RSJ IROS 2025.
\end{IEEEbiography}




\end{document}